\documentclass[conference]{IEEEtran}
\usepackage{times}

\usepackage[numbers]{natbib}
\usepackage{multicol}
\usepackage[bookmarks=true]{hyperref}

\usepackage{multirow}
\usepackage{multicol}
\usepackage{graphicx}
\usepackage{xspace}
\usepackage{xcolor}
\usepackage{caption}
\usepackage{wrapfig}
\usepackage{bbding}  
\usepackage{pifont}  

\usepackage{booktabs}
\usepackage{float}
\usepackage{amsmath}  
\usepackage{amssymb}  

\newcommand{\oursfull}[0]{3D Diffusion Policy (DP3)\xspace}
\newcommand{\ours}[0]{DP3\xspace}

\newcommand{\ourwebsite}[0]{\href{https://3d-diffusion-policy.github.io
}{3d-diffusion-policy.github.io
}\xspace}

\usepackage{colortbl}
\definecolor{ourcolor}{HTML}{99e0eb}
\definecolor{ourblue}{HTML}{27a2c3}

\definecolor{tablecolor}{HTML}{ccf2f5} 

\definecolor{tablecolor2}{HTML}{ffcdb4}
\definecolor{citecolor}{HTML}{fe7b5b}
\definecolor{grey}{rgb}{0.9, 0.9, 0.9}
\usepackage{amssymb}

\usepackage{listings}
\definecolor{gred}{rgb}{0.859,0.267,0.216}
\definecolor{ggreen}{rgb}{0.059,0.616,0.345}

\definecolor{deepblue}{HTML}{27a2c3}

\definecolor{deepred}{HTML}{fe7b5b}

\newcommand{\revise}[1]{{\color{black}#1}}

\newcommand{\dd}[2]{$#1\scriptstyle{\pm#2}$}
\newcommand{\ddbf}[2]{\cellcolor{tablecolor}$\mathbf{#1\scriptstyle{\pm#2}}$}

\newcommand{\cc}[1]{$#1$}
\newcommand{\ccbf}[1]{\cellcolor{tablecolor}$\mathbf{#1}$}

\begin{document}


\title{3D Diffusion Policy:\\
\Large{\revise{Generalizable Visuomotor Policy Learning via Simple 3D Representations}}\vspace{-0.1in}}

\author{Yanjie Ze$^{1*}$\quad Gu Zhang$^{12*}$\quad Kangning Zhang$^{12}$\quad Chenyuan Hu$^{13}$\quad Muhan Wang$^{13}$\quad Huazhe Xu$^{314}$\vspace{0.03in}\\

$^1$Shanghai Qi Zhi Institute\quad$^2$Shanghai Jiao Tong University\quad$^3$Tsinghua University, IIIS\quad$^4$Shanghai AI Lab \vspace{0.03in}\\\quad$^*$Equal contribution\vspace{0.1in}\\

\href{https://3d-diffusion-policy.github.io}{\color{deepblue}\textbf{3d-diffusion-policy.github.io}\xspace}\vspace{-0.1in}}

\twocolumn[{%
\renewcommand\twocolumn[1][]{#1}%
\maketitle
\vspace{-0.05cm}
\begin{center}
    \centering
    \captionsetup{type=figure}
     \includegraphics[width=1.0\textwidth]{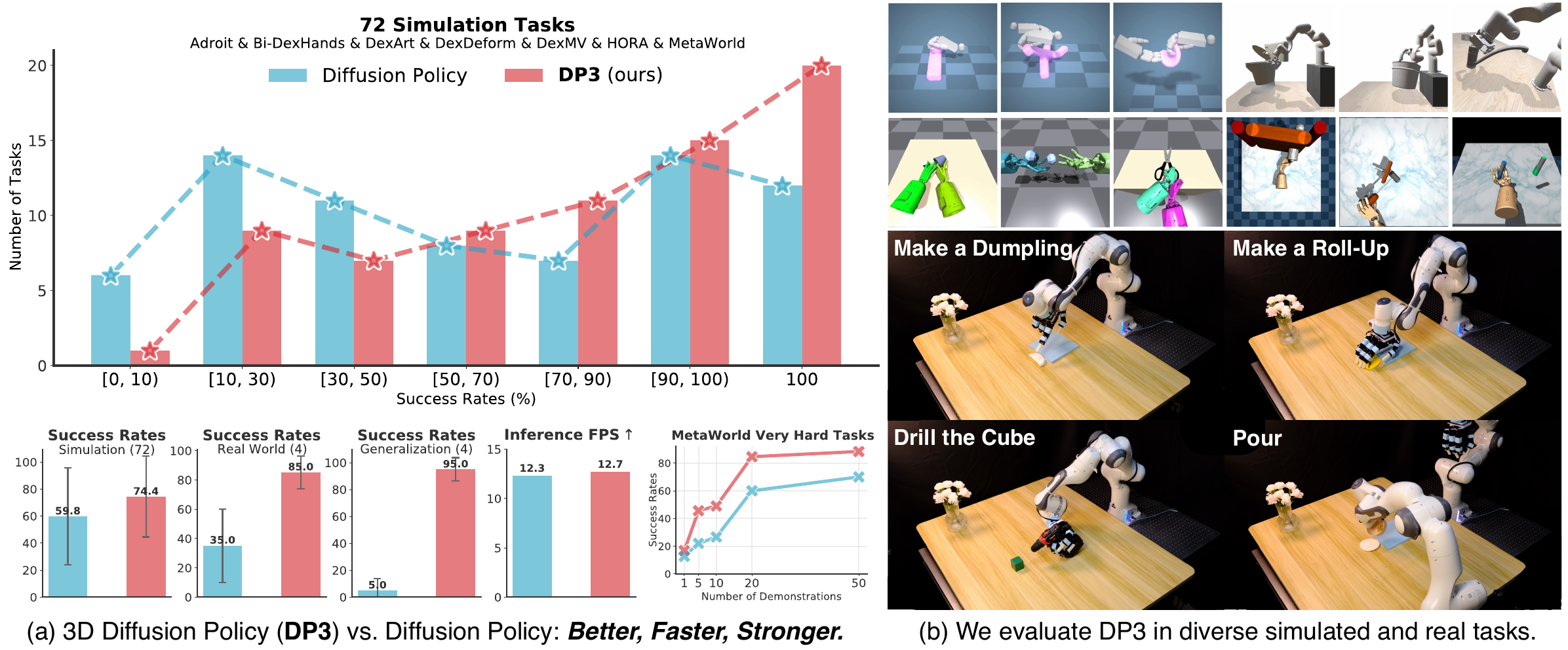}
     \vspace{-0.25in}
    \caption{\small{\textbf{\oursfull} is a visual imitation learning algorithm that marries 3D visual representations with diffusion policies, achieving surprising effectiveness in diverse simulation and real-world tasks, with a practical inference speed.}}
    \label{fig:teaser}
\end{center}
\vspace{-0.05in}
}]

\begin{abstract}

Imitation learning provides an efficient way to teach robots dexterous skills; however, learning complex skills robustly and generalizablely usually consumes large amounts of human demonstrations. To tackle this challenging problem, we present 3D Diffusion Policy (DP3), a novel visual imitation learning approach that incorporates the power of 3D visual representations into diffusion policies, a class of conditional action generative models. The core design of DP3 is the utilization of a compact 3D visual representation, extracted from sparse point clouds with an efficient point encoder. In our experiments involving 72 simulation tasks, DP3 successfully handles most tasks with just 10 demonstrations \revise{and surpasses baselines with a 24.2\% relative improvement.} In 4 real robot tasks, DP3 demonstrates precise control \revise{with a high success rate of 85\%}, given only 40 demonstrations of each task, and shows excellent generalization abilities in diverse aspects, including space, viewpoint, appearance, and instance.  \revise{Interestingly,} in real robot experiments, DP3 rarely violates safety requirements, in contrast to baseline methods which frequently do, necessitating human intervention. Our extensive evaluation highlights the critical importance of 3D representations in real-world robot learning. Code and videos are available on \ourwebsite.

\end{abstract}

\IEEEpeerreviewmaketitle

\section{Introduction}

Imitation learning provides an efficient way to teach robots a wide range of motor skills, such as grasping~\citep{wang2023mimicplay,shridhar2023peract,ze2023gnfactor}, legged locomotion~\cite{peng2020imitating_animal}, dexterous manipulation~\citep{agarwal2023functional,haldar2023fish}, humanoid loco-manipulation~\citep{seo2023humanoid}, and mobile manipulation~\citep{shafiullah2023bringing,fu2024mobile}. Visual imitation learning, which takes high-dimensional visual observations such as images or depth maps, eases the need for task-specific state estimation and thus gains the popularity~\citep{chi2023diffusion_policy,shridhar2023peract,ze2023gnfactor,florence2022ibc,hansen2022lfs}.

However, the generality of visual imitation learning comes at a cost of vast demonstrations~\citep{haldar2023fish,chi2023diffusion_policy,florence2022ibc}.
For example, the state-of-the-art method Diffusion Policy~\citep{chi2023diffusion_policy} necessitates 100 to 200 human-collected demonstrations for each real-world task. To collect the required extensive number of demonstrations, the entire data-gathering process can span several days due to its long-horizon nature and failure-prone process. 
One solution is online learning~\citep{haldar2023fish}, where the policy continues to evolve through interaction with environments and a learned reward function from expert demonstrations. Nevertheless, online learning in real-world scenarios introduces its own challenges, such as safety considerations, the necessity for automatic resetting, human intervention, and additional robot hardware costs. Therefore, how to enable (offline) imitation learning algorithms to learn robust and generalizable skills with as few demonstrations as possible is a fundamental problem, especially for practical real-world robot learning.



To tackle this challenging problem,  we introduce \textbf{\oursfull}, a  simple yet effective visual imitation learning algorithm that integrates the strengths of 3D visual representations with diffusion policies. \ours encodes sparsely sampled point clouds into a compact 3D representation using a straightforward and efficient MLP encoder. Subsequently, \ours denoises random noise into a coherent action sequence, conditioned on this compact 3D representation and the robot poses. This integration leverages not only the spatial understanding capabilities inherent in 3D modalities but also the expressiveness of diffusion models.

To comprehensively evaluate \ours, we have developed a simulation benchmark comprising 72 diverse robotic tasks from 7 domains, alongside \revise{4 real-world tasks including challenging dexterous manipulation on deformable objects.} Our extensive experiments demonstrate that although \ours is conceptually straightforward, it exhibits several notable advantages over 2D-based diffusion policies and other baselines:

\begin{enumerate}
\item \textbf{Efficiency \& Effectiveness.} \ours not only achieves superior accuracy but also does so with significantly fewer demonstrations and fewer training steps.
\item \textbf{Generalizability.} The 3D nature of \ours facilitates generalization capabilities across multiple aspects: space, viewpoint, instance, and appearance. 
\item \revise{\textbf{Safe deployment.} 
An interesting observation in our real-world experiments is that DP3 seldom gives erratic commands in real-world tasks, unlike baseline methods which often do} and exhibit unexpected behaviors, posing potential damage to the robot hardware.
\end{enumerate}

We conduct several analyses of our 3D visual representations. Intriguingly, we observed that while other baseline methods, such as BCRNN~\citep{mandlekar2021robomimic} and IBC~\citep{florence2022ibc}, benefit from the incorporation of 3D representations, they do not achieve enhancements comparable to \ours. Additionally, \ours consistently outperforms other 3D modalities, including depth and voxel representations, and surpasses other point encoders like PointNeXt~\citep{qian2022pointnext} and Point Transformer~\citep{zhao2021point_transformer}. These ablation studies highlight that the success of \ours is not just due to the usage of 3D visual representations, but also because of its careful design.

\revise{In summary, our contributions are four-fold:
\begin{enumerate}
    \item We propose 3D Diffusion Policy (DP3), an effective visuomotor policy that generalizes across diverse aspects with few demonstrations.
    \item To reduce the variance brought by benchmarks and tasks, we evaluate DP3 in a broad range of simulated and real-world tasks, showing the universality of DP3.
    \item We conduct comprehensive analyses on visual representations in DP3 and show that a simple point cloud representation is preferred over other intricate 3D representations and is better suited for diffusion policies over other policy backbones.
    \item DP3 is able to perform real-world deformable object manipulation using a dexterous hand with only 40 demonstrations, demonstrating that complex high-dimensional tasks could be handled with little human data.
\end{enumerate}
}

\ours emphasizes the power of marrying 3D representations with diffusion policies in real-world robot learning. Code is available on \url{https://github.com/YanjieZe/3D-Diffusion-Policy}.

\section{Related Work}
\subsection{Diffusion Models in Robotics}
Diffusion models, a category of generative models that progressively transform random noise into a data sample, have achieved great success in high-fidelity image generation~\citep{ho2020ddpm, song2020score, rombach2022stable_diffusion,song2020ddim}. Owing to their impressive expressiveness, diffusion models have recently been applied in robotics, including in fields such as reinforcement learning~\citep{wang2022diffusion_QL,ajay2022conditional_generative_modeling}, imitation learning~\citep{chi2023diffusion_policy, pearce2023imitating_diffusion, reuss2023goal,xian2023chaineddiffuser,sridhar2024memoryconsistent,prasad2024consistency}, reward learning~\citep{Huang2023DiffusionReward,nuti2023extracting}, grasping~\citep{wu2023graspGF,urain2023se3,simeonov2023shelving}, and motion planning~\citep{saha2023edmp,janner2022diffusion_planning}. In this work, we focus on representing visuomotor policies as conditional diffusion models, referred to as \textit{diffusion policies}, following the framework established in \citep{chi2023diffusion_policy,pearce2023imitating_diffusion}. Unlike prior methods that primarily focus on images and states as conditions, we pioneer in incorporating 3D conditioning into diffusion policies.

\subsection{Visual Imitation Learning}
\label{section: visual imitation learning}
Imitation learning offers an efficient way for robots to acquire human-like skills, typically relying on extensive observation-action pairs from expert demonstrations. Given the challenges in accurately estimating object states in the real world, visual observations such as images have emerged as a practical alternative. While 2D image-based policies~\citep{pari2021VINN,florence2022ibc,chi2023diffusion_policy,mandlekar2021robomimic,haldar2023fish, shafiullah2022bet, wang2023mimicplay,ha2023scaling} have predominated the field, the significance of 3D is increasingly recognized~\citep{shridhar2023peract,ze2023gnfactor,Ze2022rl3d,goyal2023rvt,gervet2023act3d,3d_diffuser_actor,wang2024dexcap}.

\revise{Recent 3D-based policies, including PerAct~\citep{shridhar2023peract}, GNFactor~\citep{ze2023gnfactor}, RVT~\citep{goyal2023rvt}, ACT3D~\citep{gervet2023act3d}, and NeRFuser~\citep{yan2024nerfuser}, have demonstrated notable advancements in low-dimensional control tasks. However, these works face two primary challenges: (1) \textbf{Impractical setting.} These methods convert the imitation learning problem into a prediction-and-planning paradigm using keyframe pose extraction. While effective, this formulation is less suitable for high-dimensional control tasks. (2) \textbf{Slow inference.} The intricate architectures of these methods result in slow inference speeds. For instance, PerAct~\citep{shridhar2023peract} runs at an inference speed of 2.23 FPS, making it hard to address tasks that require dense commands, such as highly dynamic environments. Another closely related work 3D Diffuser Actor~\citep{3d_diffuser_actor} runs at 1.67 FPS mainly due to the usage of attention to language tokens and the difference in the task setting\footnote{The inference speed depends on a number of factors, such as the number of camera views used, the input observation size in pixels, the number of diffusion timesteps and the temporal horizon of prediction. Since the two papers tested on different setups, the authors of 3D Diffuser Actor~\citep{3d_diffuser_actor} used DP3’s code to set it up on CALVIN, a multi-task language-conditioned setup. They equipped DP3 with attention to language tokens, identical to their model for fair comparison. They used two cameras to unproject and obtain a point cloud. They ran both DP3 and 3D Diffuser Actor on the same NVIDIA 2080 Ti GPU. Inference for 3D Diffuser Actor takes 600ms and predicts 6 action steps. Inference for DP3 takes 581ms and predicts 4 action steps. As a result, the control frequency of 3D Diffuser Actor on CALVIN is higher than DP3's.}.
Compared to this line of works, we endeavor to develop a universal and fast 3D policy capable of tackling a broader spectrum of robotic tasks, encompassing both high-dimensional and low-dimensional control tasks.}

\begin{figure*}[t]
    \centering    
\includegraphics[width=1.0\textwidth]{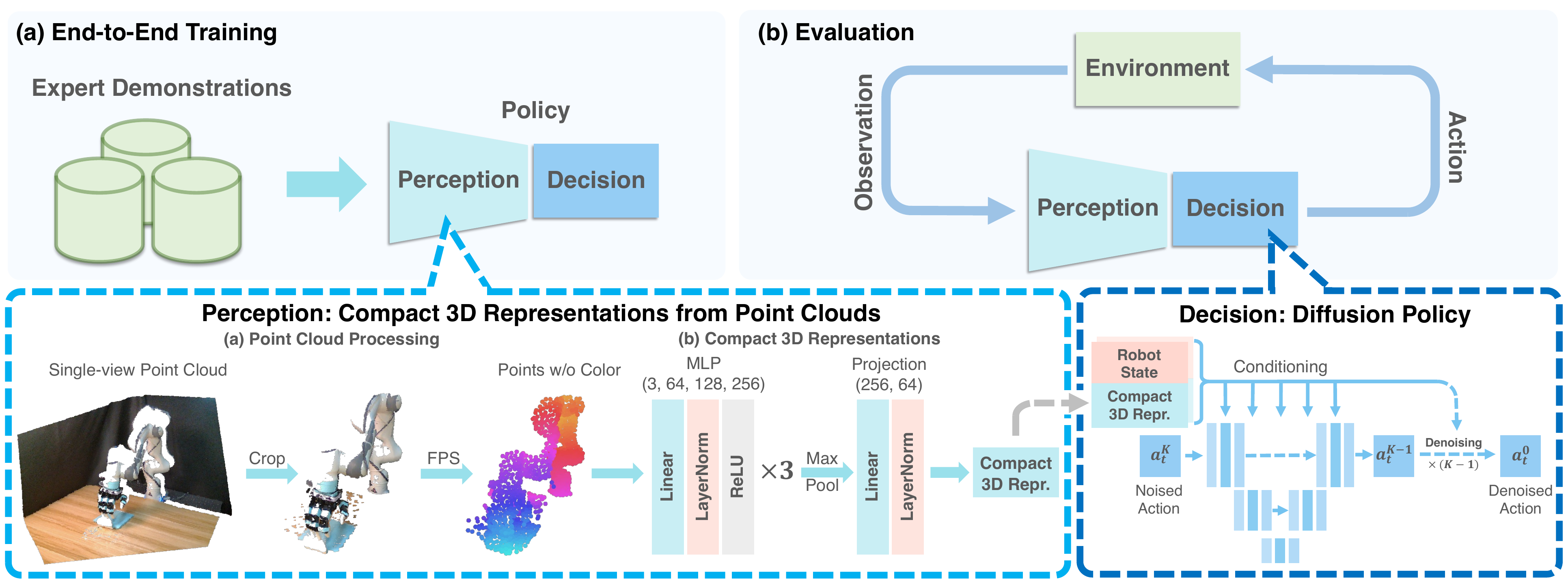}
    \caption{\textbf{Overview of 3D Diffusion Policy (DP3).} \textbf{Above:} In the training phase, DP3 simultaneously trains its perception module and decision-making process in an end-to-end manner using expert demonstrations. During evaluation, DP3 determines actions based on visual observations from the environment.
\textbf{Below:} DP3 perceives its environment through single-view point clouds. These are transformed into compact 3D representations by a lightweight MLP encoder. Subsequently, DP3 generates actions conditioning on these 3D representations and the robot's states, using a diffusion-based backbone.}
    \label{fig:method}
    \vspace{-0.2in}
\end{figure*}

\subsection{Learning Dexterous Skills} 
Achieving human-like manipulation skills in robots has been a longstanding objective pursued by robotics researchers. Reinforcement learning has been a key tool in this endeavor, enabling robots with dexterous hands to master a variety of tasks, such as pouring water~\citep{qin2022dexmv,ze2023hindex}, opening doors~\citep{rajeswaran2017dapg,hansen2023modem,chen2022bidexhands}, rotating objects~\citep{qi2023hora,yin2023rotating,yuan2023robot_synthethesia,qi2023general_hora}, reorienting objects~\citep{handa2023dextreme,chen2023visual_dex,chen2022system}, spinning pens~\citep{ma2023eureka}, grasping tools~\citep{agarwal2023functional}, executing handovers~\citep{zhang2023flexible_handover,huang2023dynamic_handover}, and building Legos~\citep{chen2023sequential}. Imitation learning offers another pathway, with approaches like DIME~\citep{arunachalam2023dime} and DexMV~\citep{qin2022dexmv} translating human hand movements into robotic actions through retargeting and enabling learning from human videos. Our work, however, diverges from these specific design-centric methods. We demonstrate that enabling the acquisition of these complex skills with minimal demonstrations could be achieved by improving the imitation learning algorithm itself.

\section{Method}
Given a \textit{small} set of expert demonstrations that contain complex robot skill trajectories, we want to learn a visuomotor policy $\pi: \mathcal{O}\mapsto\mathcal{A}$ that maps the visual observations $o\in \mathcal{O}$ to actions $a\in \mathcal{A}$, such that our robots not only reproduce the skill but also generalize beyond the training data. To this end, we introduce \textbf{\oursfull}, which mainly consists of two critical parts: (a) \textbf{Perception.} \ours perceives the environments with point cloud data and processes these visual observations with an efficient point encoder into visual features; (b) \textbf{Decision.} \ours utilizes the expressive Diffusion Policy~\citep{chi2023diffusion_policy} as the action-making backbone, which generates action sequences conditioning on our 3D visual features. An overview of \ours is in Figure~\ref{fig:method}. We will detail each part in the following sections.

\subsection{A Motivating Example}
\begin{figure}[htbp]
    \centering
    \vspace{-0.05in}
\includegraphics[width=0.5\textwidth]{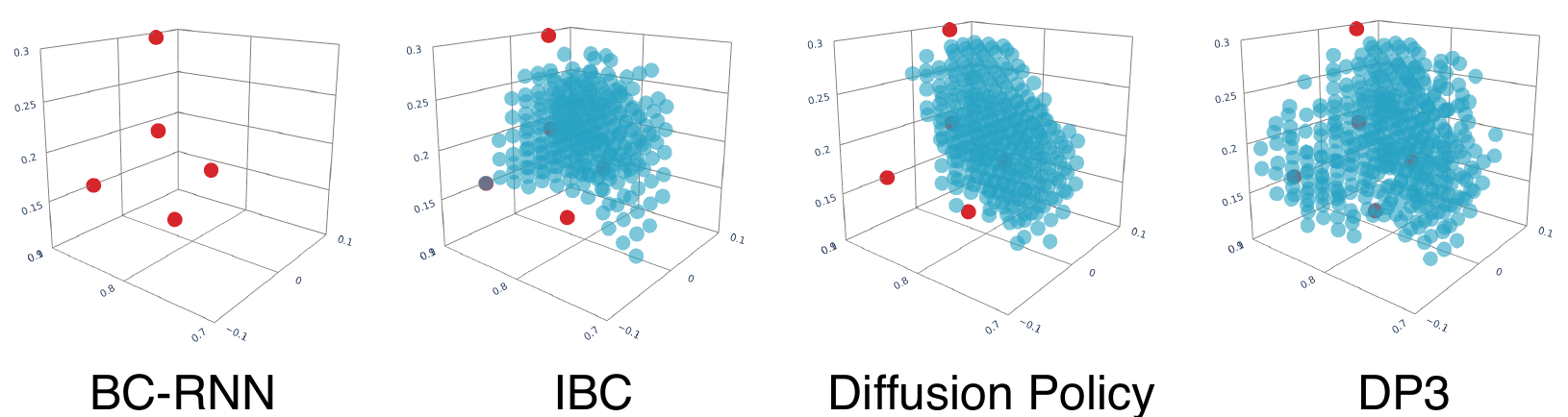}
    \caption{\textbf{Generalization in 3D space with few data.} We use MetaWorld Reach as an example task, given only 5 demonstrations (visualized by  \textcolor{deepred}{$\bullet$}). We evaluate 1000 times to cover the 3D space and visualize the \textcolor{deepblue}{$\bullet$} successful evaluation points. \textbf{\ours} learns the generalizable skill in 3D space; \textbf{Diffusion Policy} and \textbf{IBC}~\citep{florence2022ibc} only succeed in partial space; \textbf{BC-RNN}~\citep{mandlekar2021robomimic} fails to learn such a simple skill with limited data. Number of successful trials from left to right: $0$ /	$285$ / $327$ / $415$.}
    \label{fig:motivation example}
     \vspace{-0.1in}
\end{figure}

To better illustrate the generalization ability of \ours, we first give a straightforward example. We use the MetaWorld Reach task~\citep{yu2020metaworld} as our testbed. In this task, the goal is for the gripper to accurately reach a designated target point. To evaluate the effectiveness of imitation learning algorithms in not only fitting training data but also generalizing to new scenarios, we visualize the \textcolor{deepred}{$\bullet$} training points and the \textcolor{deepblue}{$\bullet$} successful evaluation points in 3D space, as shown in Figure~\ref{fig:motivation example}. We observe that with merely five training points, \ours reaches points distributed over the 3D space, while for 2D-based methods, Diffusion Policy~\citep{chi2023diffusion_policy} and IBC~\citep{florence2022ibc} learn to reach within a plane-like area, and  BCRNN~\citep{mandlekar2021robomimic} fails to cover the space. This example demonstrates the superior generalization and efficiency of \ours, particularly in scenarios where available data is limited.

\subsection{Perception}

We now detail the \textit{perception} module in \ours. \ours focuses on only utilizing a \textit{\textbf{single-view}} camera for policy learning for all the tasks, which is different from previous works~\citep{chi2023diffusion_policy,handa2023dextreme} that set up multiple cameras around robots. This is primarily chosen for its practical applicability in real-world tasks.

\textbf{Representing 3D scenes with point clouds.} The 3D scene could be represented in different ways, such as RGB-D images, point clouds, voxels~\citep{chen2023visual_dex}, implicit functions~\citep{mildenhall2021nerf}, and 3D gaussians~\citep{kerbl2023gaussian_splat}. Among them, \ours uses sparse point clouds as the 3D representation. As evidenced in our ablations (see Table~\ref{table: different 3D form}), point clouds are found to be more efficient compared to other explicit representations, such as RGB-D, depth, and voxels. 

For both simulation and the real world, we obtain depth images with size $84\times 84$ from a single camera. We then convert depth into point clouds with camera extrinsics and intrinsics. We do not use color channels for better appearance generalization.

\begin{table*}[t]
\centering
\caption{\textbf{Main simulation results.} Averaged over 72 tasks, \ours achieves $\mathbf{24.2\%}$ relative improvement compared to Diffusion Policy, with a smaller variance. Success rates for individual tasks are in Appendix~\ref{appendix: full simulation results}.}
\label{table:simulation simplified}
\vspace{-0.05in}
\resizebox{1.0\textwidth}{!}{%
\begin{tabular}{l|cccccccccc|cc}
\toprule


  & Adroit &  Bi-DexHands & DexArt & DexDeform & DexMV& HORA & MetaWorld & MetaWorld   & MetaWorld   & MetaWorld  &  \multicolumn{2}{c}{Average}\\

Algorithm $\backslash$ Task & (3) & (6) &  (4) & (6) & (2) & (1) & Easy (28) & Medium (11) & Hard (6) & Very Hard (5)& \multicolumn{2}{c}{(72)} \\

\midrule

\textbf{\ours} &  \ccbf{68.3} & \ccbf{70.2} & \ccbf{68.5} & 87.8 & \ccbf{99.5} & \ccbf{71.0} &  \ccbf{90.9} & \ccbf{61.6} & \ccbf{31.7} & \ccbf{49.0} &  \multicolumn{2}{c}{\ddbf{74.4}{29.9} ($\uparrow \mathbf{24.2}\%$)}\\

Diffusion Policy & $31.7$ & $61.3$ & $49.0$ & \ccbf{90.5} & $95.0$ & $49.0$ &  $83.6$ & $31.1$ & $9.0$ & $26.6$ & \multicolumn{2}{c}{\dd{59.8}{35.9}}\\

\bottomrule
\end{tabular}}
\vspace{-0.1in}
\end{table*}

\begin{table*}[t]
\centering
\caption{\textbf{Comparing \ours with more baselines in simulation.} We include IBC, BCRNN, and their 3D variants, termed as IBC+3D and BCRNN+3D. The 3D variants use our DP3 Encoder for a fair comparison. }
\vspace{-0.05in}
\label{table: compare with more baselines}
\resizebox{0.9\textwidth}{!}{%
\begin{tabular}{l|ccc|ccc|cccc|cc}
\toprule

    & \multicolumn{3}{c|}{Adroit} & \multicolumn{3}{c|}{MetaWorld} & \multicolumn{4}{c|}{DexArt} & \\
  Algorithm $\backslash$  Task & Hammer & Door  & Pen & Assembly & Disassemble & Stick-Push & Laptop & Faucet & Toilet & Bucket & \textbf{Average} \\

\midrule

\textbf{\ours} & \ddbf{100}{0} & \ddbf{62}{4} & \ddbf{43}{6} & \ddbf{99}{1} & \ddbf{69}{4} & \ddbf{97}{4} & \ddbf{83}{1} & \ddbf{63}{2} & \ddbf{82}{4} & \ddbf{46}{2} & \ccbf{74.4}\\

Diffusion Policy &	\dd{48}{17} & \dd{50}{5} & \dd{25}{4}	& \dd{15}{1} & \dd{43}{7} & 	\dd{63}{3} & \dd{69}{4} & \dd{23}{8}  & \dd{58}{2} & \ddbf{46}{1} & $44.0$ \\

BCRNN & \dd{0}{0} & \dd{0}{0} & \dd{9}{3} & \dd{3}{4} & \dd{32}{12} & \dd{45}{11} &  \dd{3}{3} & \dd{1}{0} & \dd{5}{5} & \dd{0}{0} & $9.8$  \\
BCRNN+3D & \dd{8}{14} & \dd{0}{0} & \dd{8}{1} & \dd{1}{5} & \dd{11}{6} & \dd{0}{0} &  \dd{29}{12} & \dd{26}{2} & \dd{38}{10} & \dd{24}{11} &  $14.5$ \\
IBC & \dd{0}{0} & \dd{0}{0} & \dd{9}{2} & \dd{0}{0} & \dd{1}{1} & \dd{16}{2} & \dd{3}{2} & \dd{7}{1} & \dd{14}{1} & \dd{0}{0} & $5.0$ \\
IBC+3D & \dd{0}{0} & \dd{0}{0} & \dd{10}{1} & \dd{18}{9} & \dd{3}{5} & \dd{50}{6} & \dd{1}{1} & \dd{7}{2} & \dd{15}{1} & \dd{0}{0} & $10.4$\\

\bottomrule
\end{tabular}}
\vspace{-0.2in}
\end{table*}

\textbf{Point cloud processing.} Since the point clouds converted from depth may contain \textit{redundant} points, such as points from the table and the ground, we crop out these points and only leave points within a bounding box. 

We further downsample points by farthest point sampling (FPS, \citep{qi2017pointnet}), which helps cover the 3D space sufficiently and reduces the randomness of point cloud sampling, compared to uniform sampling. In practice, we find downsampling $512$ or $1024$ points is sufficient for all the tasks in both simulation and the real world.

\textbf{Encoding point clouds into compact representations.} We then encode point clouds into compact 3D representations with a lightweight MLP network, as shown in Figure~\ref{fig:method}. The network, termed as \textit{DP3 Encoder}, is conceptually simple: it consists of a three-layer MLP, a max-pooling function as an order-equivariant operation to pool point cloud features, and a projection head to project the features into a compact vector. LayerNorm layers are interleaved to stabilize training~\citep{hansen2023tdmpc2}. The final 3D feature, denoted as $v$, is only $64$ dimension. As shown in our ablation studies (see Table~\ref{table: different 3D encoder}), this simple encoder could even outperform pre-trained point encoders such as PointNeXt~\citep{qian2022pointnext}, aligning with observations from \citep{hansen2022lfs}, where a properly designed small encoder is better than pre-trained large encoders in visuomotor control tasks.

\subsection{Decision}

\textbf{Conditional action generation.} The \textit{decision} module in \ours is formulated as a conditional denoising diffusion model~\citep{ho2020ddpm,chi2023diffusion_policy,pearce2023imitating_diffusion} that conditions on 3D visual features $v$ and robot poses $q$, then denoises a random Gaussian noise into actions $a$. Specifically, starting from a Gaussian noise $a^K$, the denoising network $\boldsymbol{\epsilon}_\theta$ performs $K$ iterations to gradually denoise a random noise $a^K$ into the noise-free action $a^0$,
\begin{equation}
{a}^{k-1}=\alpha_k\left(a^k-\gamma_k \boldsymbol{\epsilon}_\theta\left(a^k, k, v, q\right)\right)+\sigma_k \mathcal{N}(0, \mathbf{I})\,,
\end{equation}
where $\mathcal{N}(0, \mathbf{I})$ is Gaussian noise, $\alpha_k$, $\gamma_k$, and $\sigma_k$ are functions of $k$ and depend on the noise scheduler. This process is also called the \textit{reverse process}~\citep{ho2020ddpm}.

\textbf{Training objective.} To train the denoising network $\boldsymbol{\epsilon}_\theta$, we randomly sample a data point $a^0$ from the dataset and do a diffusion process~\citep{ho2020ddpm} on the data point to get the noise at $k$ iteration $\boldsymbol{\epsilon}^k$. The training objective is to predict the noise added to the original data,
\begin{equation}
\label{eq: loss}
\mathcal{L}=\text{MSE}\left(\boldsymbol{\epsilon}^k, \boldsymbol{\epsilon}_\theta(\bar{\alpha_k} a^0 + \bar{\beta_k} \boldsymbol{\epsilon}^k, k, v, q)\right)\,,
\end{equation}
where $\bar{\alpha_k}$ and $\bar{\beta_k}$ are noise schedule that performs one step noise adding~\citep{ho2020ddpm}.

\textbf{Implementation details.} We use the convolutional network-based diffusion policy~\citep{chi2023diffusion_policy}.  We use DDIM~\citep{song2020ddim} as the noise scheduler and use sample prediction instead of epsilon prediction for better high-dimensional action generation, with  100 timesteps at training and 10 timesteps at inference. \revise{We train 1000 epochs for MetaWorld tasks due to its simplicity} and 3000 epochs for other simulated and real-world tasks, with batch size 128 for \ours and all the baselines.

\section{Simulation Experiments}

\subsection{Experiment Setup}

\begin{figure}[htbp]
    \centering
    \vspace{-0.1in}
    \includegraphics[width=0.5\textwidth]{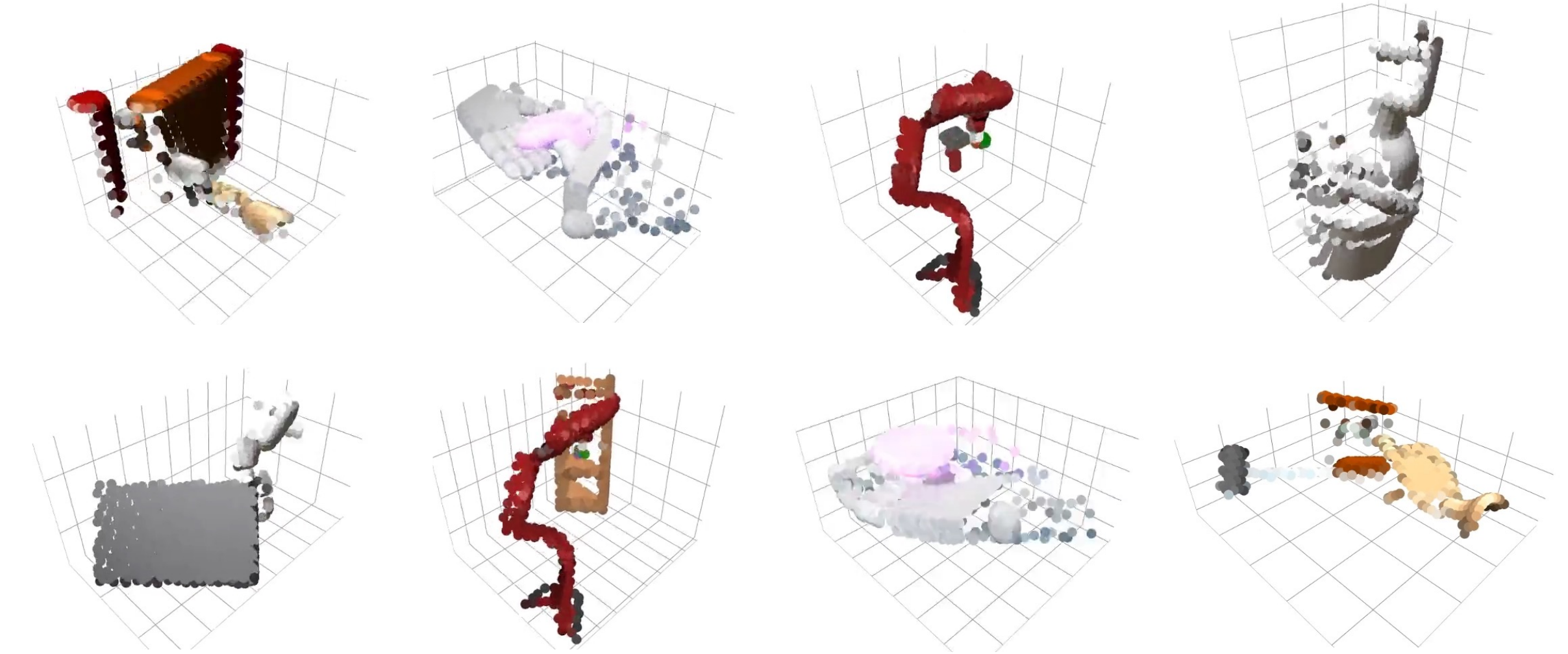}
    \caption{\textbf{3D visual observations in simulation.} We sample some simulated tasks and show the downsampled point clouds in these tasks.}
    \label{fig:sim 3d}
    \vspace{-0.15in}
\end{figure}

\textbf{Simulation benchmark.} Though the simulation environments are increasingly realistic nowadays~\citep{makoviychuk2021isaac,xiang2020sapien,todorov2012mujoco,zhu2020robosuite}, a notable gap between simulation and real-world scenarios persists~\citep{Ze2022rl3d,lei2023uni_o4,chen2023visual_dex}. This discrepancy underscores two key aspects: \textit{(a)} the importance of real robot experiments and \textit{(b)} the necessity of large-scale diverse simulation tasks for more scientific benchmarking. Therefore, for simulation experiments, we collect in total \textbf{72} tasks from \textbf{7} domains, covering diverse robotic skills. These tasks range from challenging scenarios like bi-manual manipulation~\citep{chen2022bidexhands}, deformable object manipulation~\citep{li2023dexdeform}, and articulated object manipulation~\citep{bao2023dexart}, to simpler tasks like parallel gripper manipulation~\citep{yu2020metaworld}. These tasks are built with different simulators including MuJoCo~\citep{todorov2012mujoco}, Sapien~\citep{xiang2020sapien}, IsaacGym~\citep{makoviychuk2021isaac}, and PlasticineLab~\citep{huang2021plasticinelab}, ensuring our benchmarking is not limited by the choice of simulator. Tasks in MetaWorld~\citep{yu2020metaworld} are categorized into various difficulty levels
based on \cite{seo2023mwm}. A brief overview is shown in Table~\ref{table:task summary}. \revise{The 3D observations are visualized in Figure~\ref{fig:sim 3d}.}

\textbf{Expert demonstrations.} Human-teleoperated data is used in DexDeform; \revise{Script policies are used in MetaWorld;} Trajectories for other domains are collected with agents trained by reinforcement learning (RL) algorithms, where we use VRL3~\citep{wang2022vrl3} is used for Adroit; PPO~\citep{schulman2017ppo} is used in all other domains. We generate successful trajectories with RL agents and ensure all imitation learning algorithms are using the same demonstrations. The success rates for experts are given in Appendix~\ref{appendix: full simulation results}.

\textbf{Baselines.} The primary focus of this work is to underscore the significance of the 3D modality in diffusion policies. To this end, our main baseline is the image-based diffusion policy~\citep{chi2023diffusion_policy}, simply referred to as \textit{Diffusion Policy}. Additionally, we incorporate comparisons with IBC~\citep{florence2022ibc},  BCRNN~\citep{mandlekar2021robomimic}, and their 3D variations. However, given that these algorithms showed limited effectiveness in our challenging tasks, we evaluate them on only 10 tasks (see Table~\ref{table: compare with more baselines}). We emphasize that the image and depth resolution for all 2D and 3D methods are \textbf{the same} across all experiments, ensuring a fair comparison.

\textbf{Evaluation metric.} We run 3 seeds for each experiment with seed number $0,1,2$. For each seed, we evaluate 20 episodes every 200 training epochs and then compute the average of the highest 5 success rates. We report the mean and std of success rates across 3 seeds.

\begin{table}[t]
\caption{\textbf{Task suite of \ours,} including Adroit~\citep{rajeswaran2017dapg}, Bi-DexHands~\citep{chen2022bidexhands}, DexArt~\citep{bao2023dexart}, DexDeform~\citep{li2023dexdeform}, DexMV~\citep{qin2022dexmv}, HORA~\citep{qi2023hora}, MetaWorld~\citep{yu2020metaworld}, and our real robot tasks.
ActD: the highest action dimension for the domain. \#Demo: Number of expert demonstrations used for each task in the domain. Art: articulated objects. Deform: deformable objects.}
\label{table:task summary}
\vspace{-0.05in}
\resizebox{0.5\textwidth}{!}{%
\begin{tabular}{lcccccccc}
\toprule
\toprule
\multicolumn{7}{c}{\textbf{Simulation Benchmark (72 Tasks)}} \\
\midrule
 Domain & Robo  & Object  & Simulator & ActD & \#Task & \#Demo\\
 
\midrule
\textbf{Adroit} & Shadow & Rigid/Art & MuJoCo & 28 & 3 & 10\\

\textbf{Bi-DexHands} & Shadow & Rigid/Art &  IsaacGym & 52 & 6 & 10\\

\textbf{DexArt} & Allegro & Art & Sapien & 22 & 4 & 100 \\

\textbf{DexDeform} & Shadow & Deform & PlasticineLab & 52 & 6 & 10\\

\textbf{DexMV} & Shadow  & Rigid/Fluid & Sapien & 30 & 2 & 10\\

\textbf{HORA} & Allegro  & Rigid & IsaacGym & 16 & 1 & 100\\

\textbf{MetaWorld} & Gripper & Rigid/Art & MuJoCo & 4 & 50 & 10 \\

\toprule


\end{tabular}
}
\resizebox{0.5\textwidth}{!}{%
\begin{tabular}{lcccclcccc}
\multicolumn{6}{c}{\textbf{Real Robot Benchmark (4 Tasks) }} \\
\midrule

Task & Robo & Object &  ActD & \#Demo & Description\\

\midrule
\textbf{Roll-Up} & Allegro & Deform & 22 & 40 & Wrap plasticine to make a roll-up\\

\textbf{Dumpling} & Allegro & Deform & 22 & 40 & Wrap plasticine and pinch with fingers\\

\textbf{Drill} & Allegro & Rigid & 22 & 40 & Grasp the drill and touch the cube\\


\textbf{Pour} & Gripper & Rigid & 7 & 40 & Pick the bowl, pour, and place \\
\bottomrule
\bottomrule
\end{tabular}
}
\vspace{-0.2in}
 \end{table}

\begin{figure*}[t]
    \centering
    \includegraphics[width=0.96\textwidth]{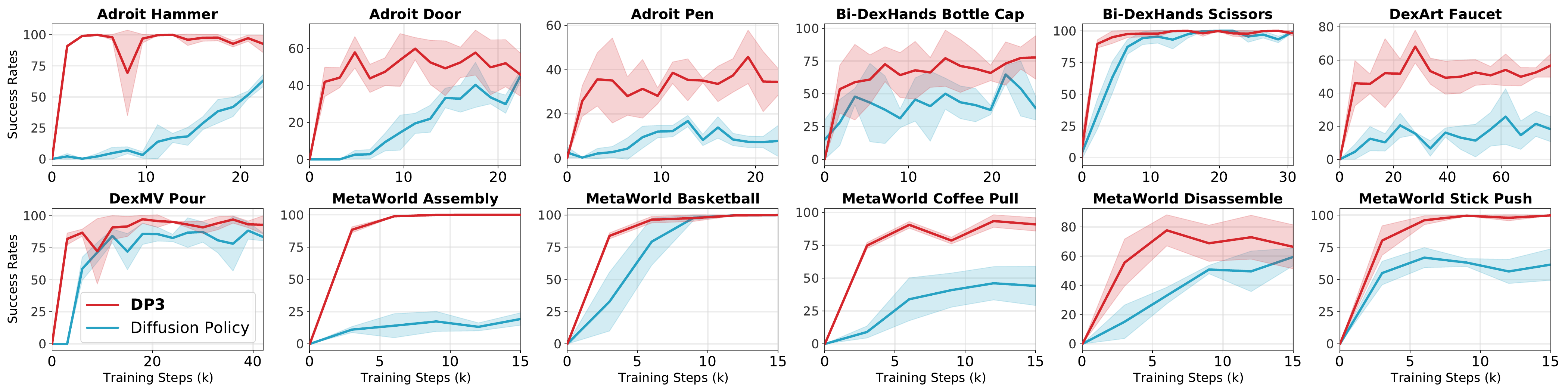}
    \vspace{-0.1in}
    \caption{\textbf{Learning efficiency.} We sample 12 simulation tasks and show the learning curves of \ours and Diffusion Policy. \ours demonstrates a rapid convergence towards high accuracy. In contrast, Diffusion Policy exhibits a slower learning progress and achieves notably lower convergence in most tasks.}
    \label{fig:learning efficiency}
    \vspace{-0.15in}
\end{figure*}
\begin{figure*}[t]
    \centering
\includegraphics[width=0.96\textwidth]{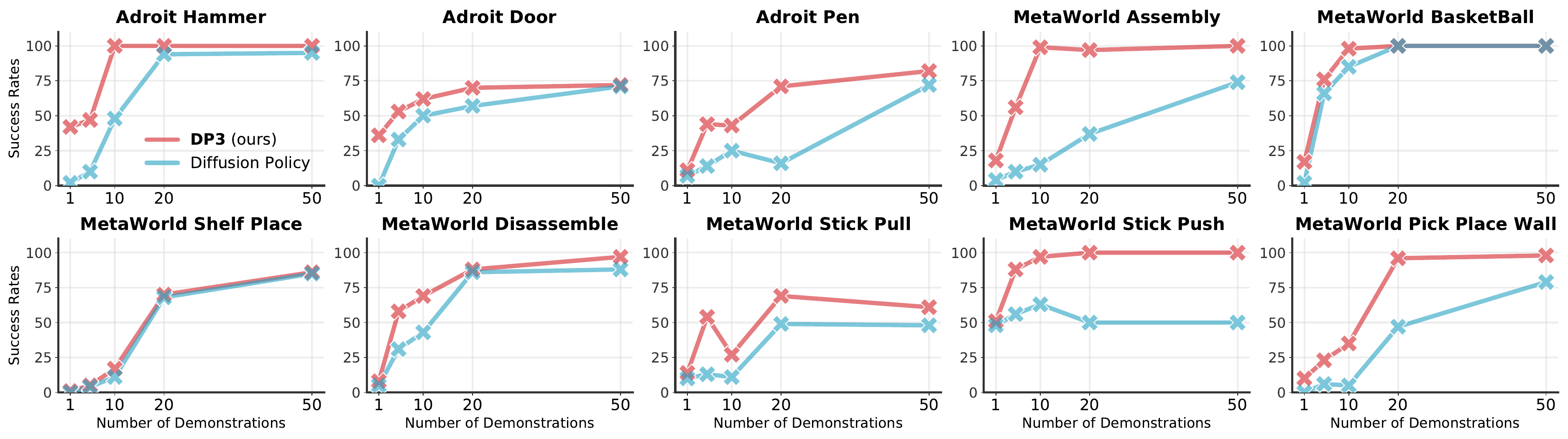}
\vspace{-0.1in}
    \caption{\textbf{Efficient scaling with demonstrations.} We sample 10 simulation tasks and train \ours and Diffusion Policy with an increasing number of demonstrations. \revise{
    \ours addresses all these tasks well and generally improves the accuracy with more demonstrations. Diffusion Policy also scales well on some tasks while still falling short of accuracy.}}
    \label{fig:demo scaling}
    \vspace{-0.25in}
\end{figure*}

\subsection{Efficiency and Effectiveness}
\ours shows surprising efficiency across diverse tasks, mainly reflected in the following three perspectives: 
\begin{enumerate}
    \item \textbf{High accuracy.} Summarized results are in Figure~\ref{fig:teaser}(a) and results for each domain are in Table~\ref{table:simulation simplified}. 
    We observe that DP3 achieves a success rate exceeding 90\% in nearly 30 tasks, whereas Diffusion Policy does in fewer than 15 tasks. Additionally, DP3 did not record any task with a success rate below 10\%, in contrast to Diffusion Policy, which had more than 10 tasks below 10\%. Note that most of the tasks are only trained with \textbf{10} demonstrations.
    \item \textbf{Learning efficiency.}
    While we train all the algorithms for 3000 epochs to guarantee convergence, we observe that \ours typically reaches convergence within approximately 500 epochs across all tasks, as illustrated in Figure~\ref{fig:learning efficiency}. In contrast, Diffusion Policy tends to converge at a much slower pace or converge into sub-optimal results.
    \item \textbf{Efficient scaling with demonstrations.} As shown in Figure~\ref{fig:demo scaling}, we find that in Adroit tasks, both \ours and Diffusion Policy perform reasonably while \ours achieves a comparable accuracy with fewer demonstrations. \revise{For some MetaWorld tasks above the \textit{easy} level such as Assembly and Disassemble, DP3 could achieve higher accuracy when demonstrations are sufficient.} This underscores that the 3D modality is not just beneficial but essential for certain manipulation tasks.
    \item \textbf{Competitive inference speed.} As depicted in Figure~\ref{fig:teaser}, \ours achieves an inference speed marginally surpassing Diffusion Policy. Contrary to the prevailing assumption that 3D-based policies are slower~\citep{shridhar2023peract,ze2023gnfactor,xian2023chaineddiffuser}, \ours manages to achieve efficient inference speeds, primarily attributed to the utilization of sparse point clouds and compact 3D representations.
\end{enumerate}

\subsection{Ablations}
We select 6 tasks to conduct more ablation studies: Hammer (\textbf{H}), Door (\textbf{D}), Pen (\textbf{P}) from Adroit and \revise{Assembly (\textbf{A}), Disassemble (\textbf{DA}),  Stick-Push (\textbf{SP})} from MetaWorld. These tasks include both high-dimensional and low-dimensional control tasks, and each task only uses 10 demonstrations. We use the abbreviations of these tasks in the tables for simplicity.


\textbf{Choice of 3D representations.} 
In \ours, we deliberately select point clouds to represent the 3D scene. To compare different choices of 3D representations,  we implement other 3D representations, including RGB-D, depth, and voxel. \revise{We also compare with oracle states, which include object states, target goals, and robot velocity besides robot poses.} The RGB-D and depth images are processed using the same image encoder as Diffusion Policy, while voxel representations employ the VoxelCNN, as implemented in \citep{chen2023visual_dex}. As demonstrated in Table~\ref{table: different 3D form}, these alternative 3D representations fall short of \ours. 
\revise{We note that RGB-D and depth images are close and not comparable to point clouds, indicating that the proper usage of depth information is essential. Additionally, we observe that point clouds and oracle states are very competitive, showing that point clouds might help learn an optimal policy from demonstrations.}

\begin{table}[htbp]
\centering
\caption{\textbf{Ablation on 3D representations.} We replace the visual observation and the corresponding encoder in \ours to evaluate different 3D representations.}
\vspace{-0.05in}
\label{table: different 3D form}
\resizebox{0.49\textwidth}{!}{%
\begin{tabular}{l|cccccc|ccc}
\toprule

  Repr. & H & D  & P & A & DA & SP & \textbf{Average} \\

\midrule
\revise{Oracle State} & \dd{99}{2} & \dd{61}{2} & \ddbf{44}{3} & \dd{94}{1} & \ddbf{72}{7} & \dd{91}{8} & $76.8$ \\

\textbf{Point cloud} & \ddbf{100}{0} & \ddbf{62}{4} & \dd{43}{6} & \ddbf{99}{1} & \dd{69}{4} & \ddbf{97}{4} & \ccbf{78.3} \\

Image & \dd{48}{17} & \dd{50}{5} & \dd{25}{4} & \dd{15}{1} & \dd{43}{7} & \dd{63}{3} & $40.7$ \\

Depth & \dd{39}{15} & \dd{49}{1} & \dd{12}{3} & \dd{15}{4}  & \dd{15}{2} & \dd{62}{3} & $32.0$ \\

RGB-D & \dd{57}{14} & \dd{47}{5} & \dd{14}{2} & \dd{15}{3} & \dd{14}{1} & \dd{61}{3} & $34.7$\\

Voxel & \dd{54}{5} & \dd{33}{3} & \dd{18}{2} & \dd{10}{2} & \dd{17}{1} & \dd{62}{6} & $32.3$\\

\bottomrule
\end{tabular}}
\vspace{-0.1in}
\end{table}

\textbf{Choice of point cloud encoders.} We compare DP3 Encoder with other widely used point encoders, including PointNet~\citep{qi2017pointnet},  PointNet++~\citep{qi2017pointnet++}, PointNeXt~\citep{qian2022pointnext}, and Point Transformer~\citep{zhao2021point_transformer}. We also include the pre-trained models of PointNet++ and PointNeXt. Surprisingly, we find that none of these complex models and the pre-trained ones are competitive to DP3 Encoder, as shown in Table~\ref{table: different 3D encoder}.


\begin{table}[htbp]
\centering
\caption{\textbf{Ablation on point cloud encoders.} We replace DP3 Encoder with other widely used encoders, including PointNet~\citep{qi2017pointnet}, PointNet++~\citep{qi2017pointnet++}, PointNeXt~\citep{qian2022pointnext}, and Point Transformer~\citep{zhao2021point_transformer}. We also include the pre-trained encoders.}
\label{table: different 3D encoder}
\vspace{-0.05in}
\resizebox{0.49\textwidth}{!}{%
\begin{tabular}{l|cccccc|ccc}
\toprule

  Encoders & H & D  & P & A & DA & SP & \textbf{Average} \\

\midrule

\textbf{DP3 Encoder} & \ddbf{100}{0} & \ddbf{62}{4} & \ddbf{43}{6} & \ddbf{99}{1} & \ddbf{69}{4} & \ddbf{97}{4} & \ccbf{78.3} \\

PointNet & \dd{46}{8} & \dd{34}{8} & \dd{14}{4} & \dd{0}{0}& \dd{0}{0}& \dd{0}{0} &  $15.7$\\

PointNet++ & \dd{0}{0} & \dd{0}{0} & \dd{13}{3} & \dd{0}{0} & \dd{0}{0}& \dd{0}{0} & $2.2$ \\

PointNeXt & \dd{0}{0} & \dd{0}{0} & \dd{14}{3} & \dd{0}{0} & \dd{0}{0}& \dd{0}{0}  & $2.3$\\

Point Transformer & \dd{0}{0} & \dd{0}{0} & \dd{6}{5}& \dd{0}{0}& \dd{0}{0}& \dd{0}{0} & $1.0$ \\

PointNet++ (pre-trained) & \dd{5}{9} & \dd{19}{12} & \dd{17}{6}  & \dd{0}{0} & \dd{0}{0} & \dd{0}{0} & $6.8$ \\

PointNeXt (pre-trained) & \dd{0}{0} & \dd{36}{13} & \dd{17}{6} & \dd{0}{0} &\dd{0}{0}  &\dd{0}{0} & $8.8$\\

\bottomrule
\end{tabular}}
\vspace{-0.1in}
\end{table}

\textbf{Gradually modifying a PointNet.} To elucidate the performance disparity between DP3 Encoder and a commonly used point cloud encoder, \textit{e.g.}, PointNet, we gradually modify a PointNet to make it aligned with a DP3 Encoder. Through extensive experiments shown in Table~\ref{table: modify PointNet}, we identify that the T-Net and BatchNorm layers in PointNet are primary inhibitors to its efficiency.  By omitting these two elements, PointNet attains an average success rate of $72.3$, competitive to $78.3$ achieved by our DP3 Encoder.One plausible explanation for the T-Net is that our control tasks use the fixed camera and  do not require feature transformations from the T-Net. \revise{Further replacing high-dimensional features with a lower-dimensional one would not hurt the performance much ($72.5\rightarrow 72.3$) but increase the speed.} We would explore the reason for the failures of other encoders in the future.
\begin{table}[htbp]
\centering
\caption{\textbf{Gradually modifying a PointNet to a DP3-style encoder.} Conv: use convolutional layers or linear layers. w/ T-Net: with or without T-Net. w/ BN: with or without BacthNorm layers. 1024 Dim: set feature dimensions before the projection layer to be 1024 or 256. Average success rates for 6 ablation tasks are reported.}
\label{table: modify PointNet}
\vspace{-0.05in}
















\resizebox{0.48\textwidth}{!}{%
\begin{tabular}{c|cccc|ccccc}
\toprule

  Encoders & Conv& w/ T-Net & w/ BN & 1024 Dim& \textbf{Average} \\
  \midrule
PointNet & \textcolor{ggreen}{\Checkmark} & \textcolor{ggreen}{\Checkmark} & \textcolor{ggreen}{\Checkmark} & \textcolor{ggreen}{\Checkmark} & $15.7$ \\

 & \textcolor{gred}{\XSolidBrush} & \textcolor{ggreen}{\Checkmark} & \textcolor{ggreen}{\Checkmark} & \textcolor{ggreen}{\Checkmark} & $15.7$ \\

 & \textcolor{ggreen}{\Checkmark} & \textcolor{gred}{\XSolidBrush} & \textcolor{ggreen}{\Checkmark} & \textcolor{ggreen}{\Checkmark} &  $16.0$\\

  & \textcolor{gred}{\XSolidBrush} & \textcolor{gred}{\XSolidBrush}  & \textcolor{ggreen}{\Checkmark} & \textcolor{ggreen}{\Checkmark} & $26.0$ \\

& \textcolor{gred}{\XSolidBrush} & \textcolor{ggreen}{\Checkmark} & \textcolor{ggreen}{\Checkmark} & \textcolor{gred}{\XSolidBrush} & $18.2$\\

  Turnaroud! & \textcolor{ggreen}{\Checkmark}  & \textcolor{gred}{\XSolidBrush} & 
    \textcolor{gred}{\XSolidBrush} &\textcolor{ggreen}{\Checkmark} &  $\mathbf{72.5}$ \\

  & \textcolor{gred}{\XSolidBrush} & \textcolor{gred}{\XSolidBrush}  & \textcolor{ggreen}{\Checkmark} & \textcolor{gred}{\XSolidBrush} & $19.8$\\

& \textcolor{gred}{\XSolidBrush} & \textcolor{ggreen}{\Checkmark} & \textcolor{gred}{\XSolidBrush} & \textcolor{gred}{\XSolidBrush} & $26.8$\\

    & \textcolor{gred}{\XSolidBrush} & \textcolor{gred}{\XSolidBrush}  & \textcolor{gred}{\XSolidBrush} & \textcolor{gred}{\XSolidBrush} & $\mathbf{72.3}$\\

\bottomrule
\end{tabular}}
\vspace{-0.1in}
\end{table}

\begin{figure}[htbp]
    \centering
    \vspace{-0.1in}
\includegraphics[width=0.48\textwidth]{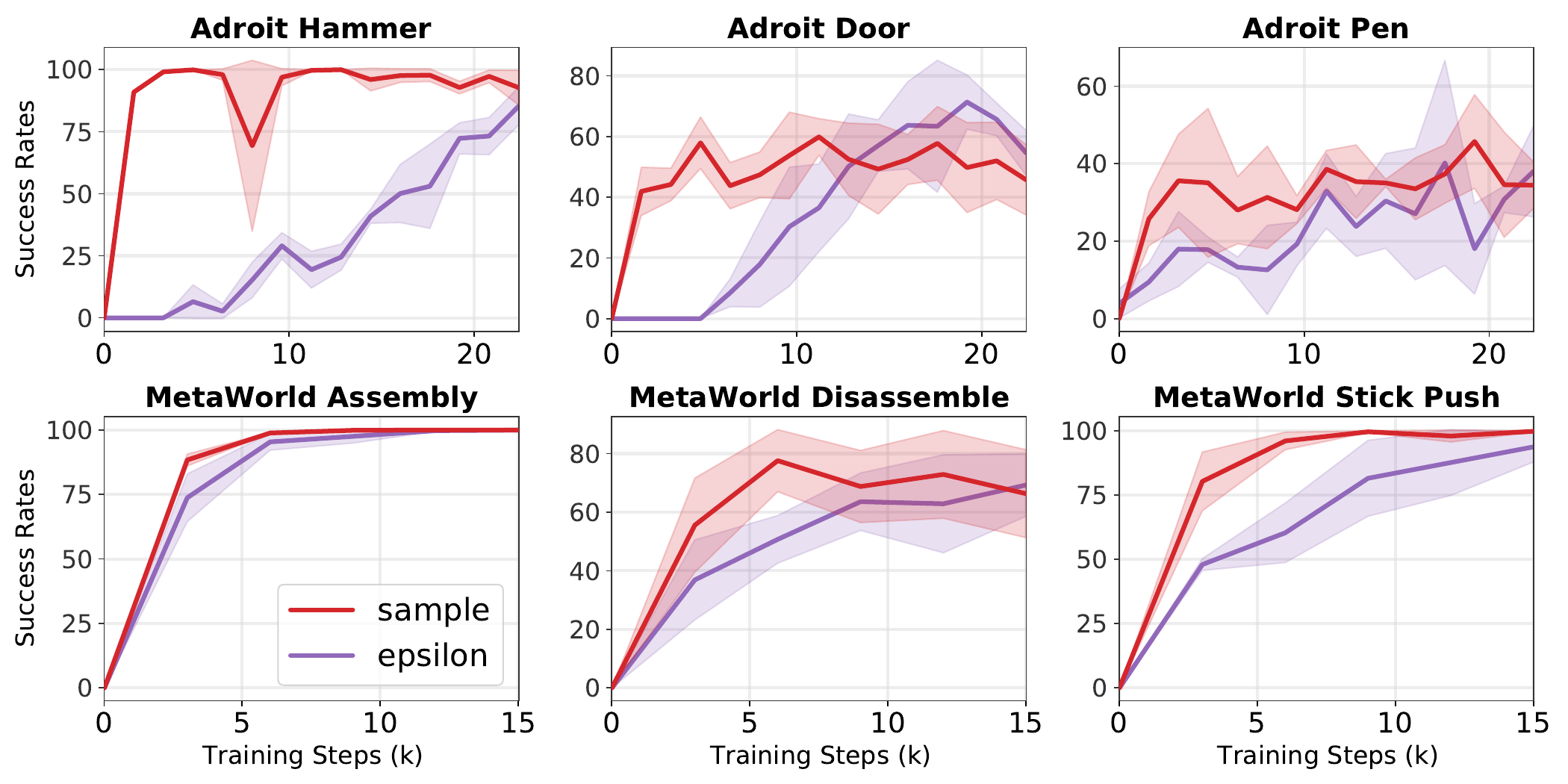}\vspace{-0.1in}
    \caption{\textbf{Learning curves of \ours with sample prediction and epsilon prediction.} With sample prediction, \ours generally converges faster, while epsilon prediction is also competitive.}
    \label{fig:learning curve sample and epsilon}
    \vspace{-0.1in}
\end{figure}

\textbf{Design choices in \ours.} Besides the 3D representations, the effectiveness of \ours is contributed by several small design choices, as shown in Table~\ref{table: ablation component}. (a) Cropping point clouds helps largely improve accuracy; (b) Incorporating LayerNorm layers could help stabilize training across different tasks~\citep{hansen2023tdmpc2,ba2016layer_norm}; (c) Sample prediction in the noise sampler brings faster convergence, also shown in Figure~\ref{fig:learning curve sample and epsilon}; (d) The projection head in DP3 Encoder accelerates the inference by projecting features to the lower dimension, without hurting accuracy; (e) Removing color channels ensures robust appearance generalization; (f) In low-dimensional control tasks, DPM-solver++~\citep{lu2022dpm_solver} as the noise sampler is competitive to DDIM, while DPM-solver++ could not handle high-dimensional control tasks well.

\begin{table}[htbp]
\centering
\caption{\textbf{Ablation on design choices in \ours.} Most of the design choices would not affect the accuracy but bring other benefits such as appearance generalization by removing color.}
\label{table: ablation component}
\vspace{-0.05in}
\resizebox{0.49\textwidth}{!}{%
\begin{tabular}{l|cccccc|ccc}
\toprule


 Designs & H & D  & P & A & DA & SP & \textbf{Average} \\

\midrule

\textbf{DP3}& \ddbf{100}{0} & \dd{62}{4} & \dd{43}{6} & \ddbf{99}{1} & \dd{69}{4} & \dd{97}{4} & \ccbf{78.3}  \\

w/o cropping & \dd{98}{1} & \ddbf{69}{3} & \dd{14}{1} & \dd{19}{9} & \dd{32}{6} & \dd{40}{2} &  $45.3$ \\

w/o LayerNorm & \ddbf{100}{0} & \dd{56}{4} & \dd{44}{3} & \dd{96}{2} & \dd{51}{3} & \dd{91}{5} &  $73.0$ \\

w/o sample pred &  \dd{68}{3} & 	\dd{67}{8} &	\dd{37}{12} & \dd{96}{2} & \dd{58}{9} & \dd{76}{9} & $67.0$ \\

w/o projection & \ddbf{100}{0} & \dd{61}{2} & \ddbf{47}{3} & \ddbf{99}{1} & \dd{60}{8} & \ddbf{99}{2} & $77.7$\\

w/ color & \ddbf{100}{1}	& \dd{67}{3}	& \dd{46}{4} & \dd{76}{8} & \ddbf{75}{5} & \dd{68}{3} & $72.0$ \\

DDIM$\rightarrow$DPM-solver++ & \dd{12}{4} & \dd{9}{5} & \dd{26}{5} & \dd{93}{3} & \dd{58}{6} & \dd{92}{14} & $48.3$  \\

\bottomrule
\end{tabular}}
\vspace{-0.05in}
\end{table}

\section{Real World Experiments}

\subsection{Experiment Setup}

\begin{figure*}[t]
    \centering
\includegraphics[width=0.99\textwidth]{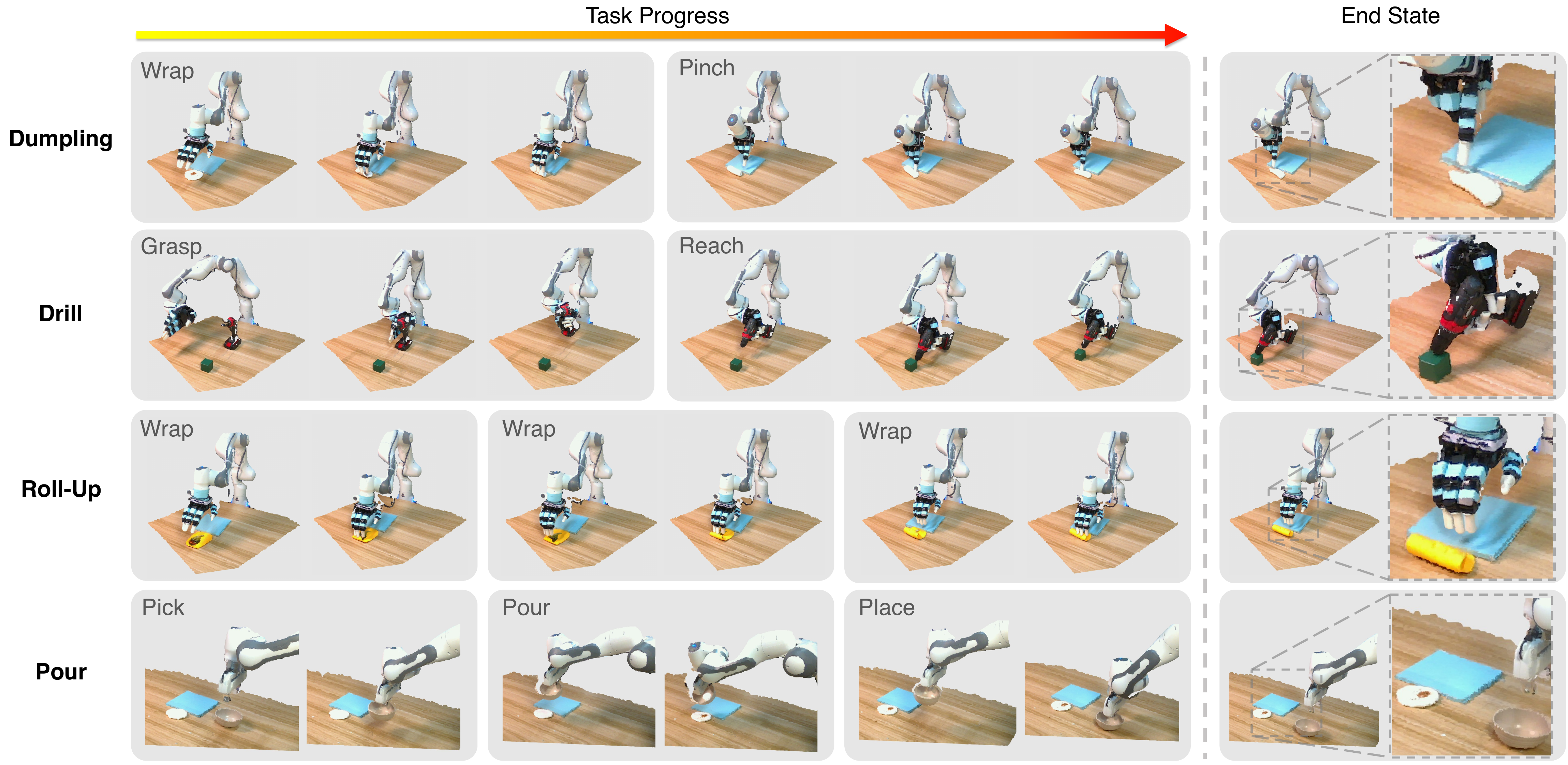}
\vspace{-0.08in}
    \caption{\textbf{Our real robot benchmark} consists of 4 challenging tasks. The Allegro hand is required to make a \textbf{\textit{Dumpling}}, \textbf{\textit{Drill}} the cube, and make a \textbf{\textit{Roll-Up}}. The gripper is required to \textbf{\textit{Pour}} dried meat floss in the bowl. Each task contains multiple stages. We visualize the point clouds of the collected trajectories.}
    \label{fig:real robot task visualization}
    \vspace{-0.17in}
\end{figure*}

\textbf{Real robot benchmark.} \ours is evaluated across 4 tasks on 2 different robots, including an Allegro hand and a gripper. We use one RealSense L515 camera to obtain real-world visual observations. All the tasks are visualized in Figure~\ref{fig:real robot task visualization} and summarized in Table~\ref{table:task summary}. \revise{Our real-world setup and everyday objects used in our tasks are shown in Figure~\ref{fig:real world setup}.} We now briefly describe our tasks:

\begin{figure}[t]
    \centering
\includegraphics[width=0.49\textwidth]{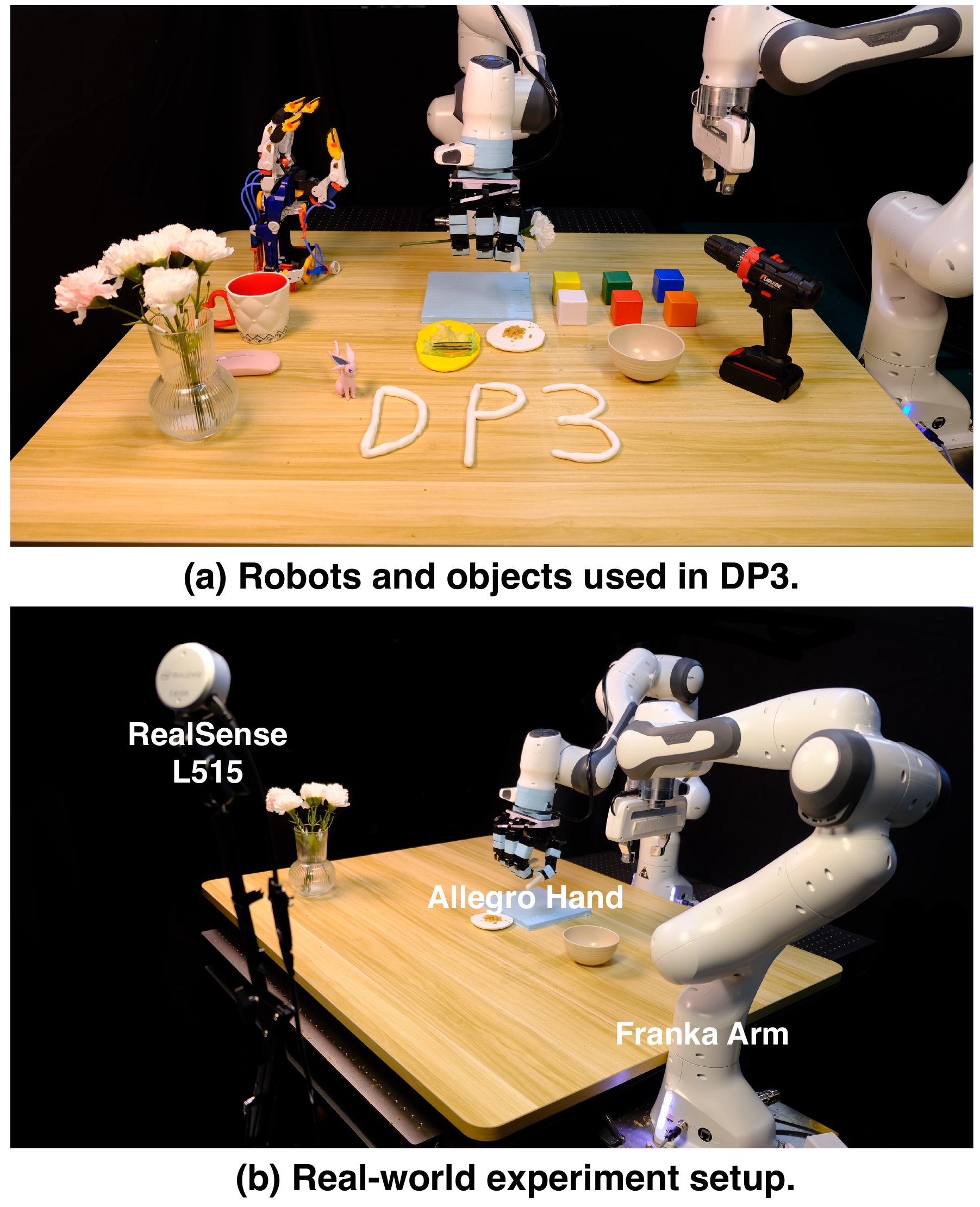}
\vspace{-0.25in}
    \caption{\revise{\textbf{(a) Our robots and objects. (b) Our real-world experiment setup.} We use an Allegro hand and a gripper based on Franka arms and include diverse everyday objects in our manipulation tasks. A RealSense L515 camera is applied to capture visual observations.}}
    \label{fig:real world setup}
    \vspace{-0.25in}
\end{figure}

\begin{figure}[htbp]
    \centering
\includegraphics[width=0.5\textwidth]{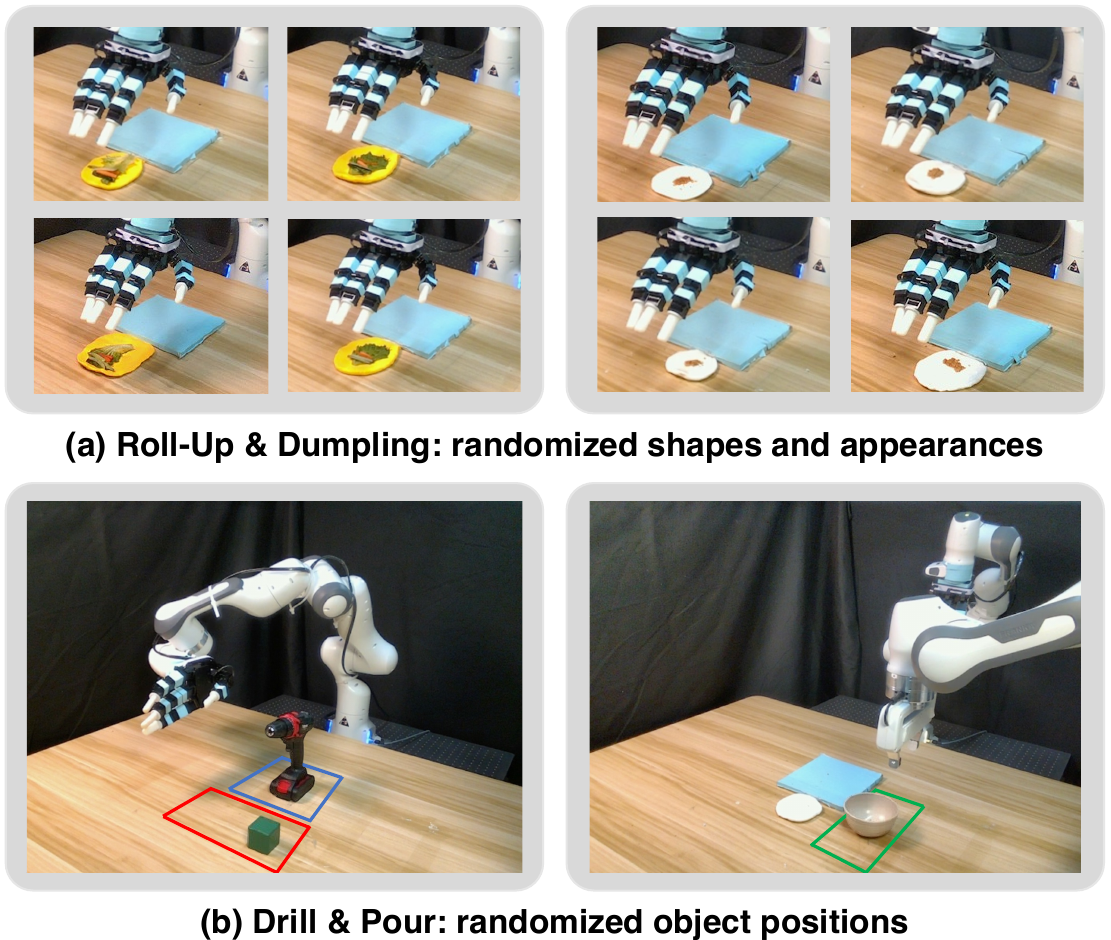}
\vspace{-0.1in}
    \caption{\revise{\textbf{Randomization in collected demonstrations for real-world tasks.} \textbf{Roll-Up}: The shape of the plasticine and the vegetables on it varies in each trajectory. \textbf{Dumpling}: The shape of the plasticine and the distribution of the meat floss on it are different in each trajectory. \textbf{Drill}: The red and blue rectangles respectively mark the range of positions where the cube and drill can be placed. \textbf{Pour}: The green rectangle marks the range of positions of the bowl.}}
    \label{fig:real task randomness}
    \vspace{-0.25in}
\end{figure}

\begin{enumerate}
    \item \textbf{Roll-Up.} The Allegro hand wraps the plasticine multiple times to make a roll-up. 
    \item \textbf{Dumpling.} The Allegro hand first wraps the plasticine and then pinchs it to make dumpling pleats. 
    \item \textbf{Drill.} The Allegro hand grasps the drill up and moves towards the green cube to touch the cube with the drill.
\item \textbf{Pour.} The gripper grasps the bowl, moves towards the plasticine, pours out the dried meat floss in the bowl, and places the bowl on the table.

\end{enumerate}
\revise{The randomization in each task is shown in Figure~\ref{fig:real task randomness}. For Roll-Up and Dumpling, the plasticine's shape and the appearance of the objects placed upon the plasticine are randomized. For Drill and Pour, the variations come from the random positions of the cube, drill, and bowl.}

Notably, our tasks using the multi-finger hand are carefully designed to show its advantage over the parallel gripper: In Roll-Up and Dumpling, robots could wrap plasticine without requiring extra tools, unlike RoboCook~\citep{shi2023robocook}; In Drill, the drill in the real world is large and heavy, which is quite difficult for the gripper to grasp.

\textbf{Expert demonstrations} are collected by human teleoperation. The Franka arm and the gripper are teleoperated by the keyboard. The Allegro hand is teleoperated with human hands by vision-based retargeting~\citep{qin2023anyteleop,handa2020dexpilot}. Since our tasks contain more than one stage and include complex multi-finger robots and deformable objects, making the process of demonstration collection very time-consuming, we only provide 40 demonstrations for each task.

\begin{figure}[t]
    \centering
    \includegraphics[width=0.5\textwidth]{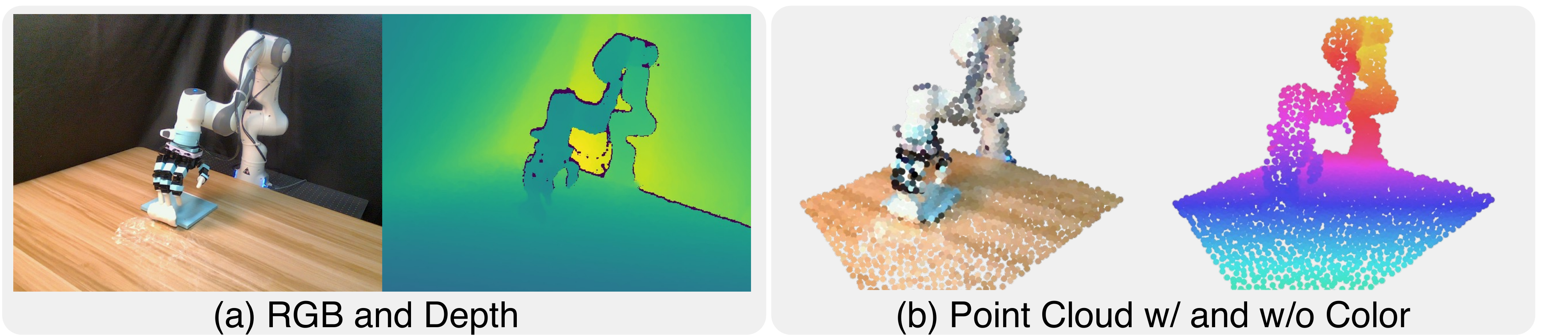}
    \vspace{-0.22in}
    \caption{\textbf{Different vision modalities in the real world,} include images, depths, and point clouds. }
    \label{fig:different 3d representations}
    \vspace{-0.12in}
\end{figure}

\textbf{Baselines.} Based on our simulation experiments, image-based and depth-based diffusion policies are still powerful, thus we select them as baselines for real-world experiments. Different vision modalities are visualized in Figure~\ref{fig:different 3d representations}.

\subsection{Effectiveness}
Results for our real robot tasks are given in Table~\ref{table: main real robot}. Consistent with our simulation findings, we observe in real-world experiments that \ours could handle all tasks with high success rates, given only \textit{40} demonstrations. Interestingly, we also observe that while both image-based and depth-based diffusion policies have comparatively low average accuracies, they exhibit distinct strengths in specific tasks. For instance, the image-based diffusion policy excels in the Drill task but fails entirely in Roll-Up. In contrast, the depth-based policy achieves a notable success rate of $40\%$ in Roll-Up.

\begin{table}[htbp]
\centering
\vspace{-0.1in}
\caption{\textbf{Main results for real robot experiments.} Each task is evaluated with 10 trials.}
\label{table: main real robot}
\vspace{-0.05in}
\resizebox{0.49\textwidth}{!}{%
\begin{tabular}{l|cccc|ccccc}
\toprule

  Real Robot & Roll-Up & Dumpling  & Drill & Pour & \textbf{Average} \\

\midrule

Diffusion Policy & \cc{0} & \cc{30} & \cc{70} & \cc{40} & \dd{35.0}{25.0} \\
Diffusion Policy (Depth) & \cc{40} & \cc{20}  &\cc{10} & \cc{10} & \dd{20.0}{12.2} \\
\textbf{\ours} & \ccbf{90} & \ccbf{70} & \ccbf{80} & \ccbf{100} &  \ddbf{85.0}{11.2}\\

\bottomrule
\end{tabular}}
\vspace{-0.2in}
\end{table}

\subsection{Generalization}
Besides the effectiveness in handling all tasks, \ours show strong generalization abilities in the real world. We categorize the generalization abilities of \ours into 4 aspects and detail each aspect as follows.

\textbf{Spatial generalization.} As illustrated in our motivating example, \ours could better extrapolate in 3D space. We demonstrate this property in the real world, as shown in Table~\ref{table: spatial generalization}. We find that baselines fail to generalize to all test positions while \ours succeed in 4 out of 5 trials.

\begin{table}[htbp]
\centering
\vspace{-0.1in}
\caption{\textbf{Spatial generalization on Pour.} We place the bowl at 5 different positions that are unseen in the training data. Each position is evaluated with one trial.}
\label{table: spatial generalization}
\vspace{-0.05in}
\includegraphics[width=0.40\textwidth]{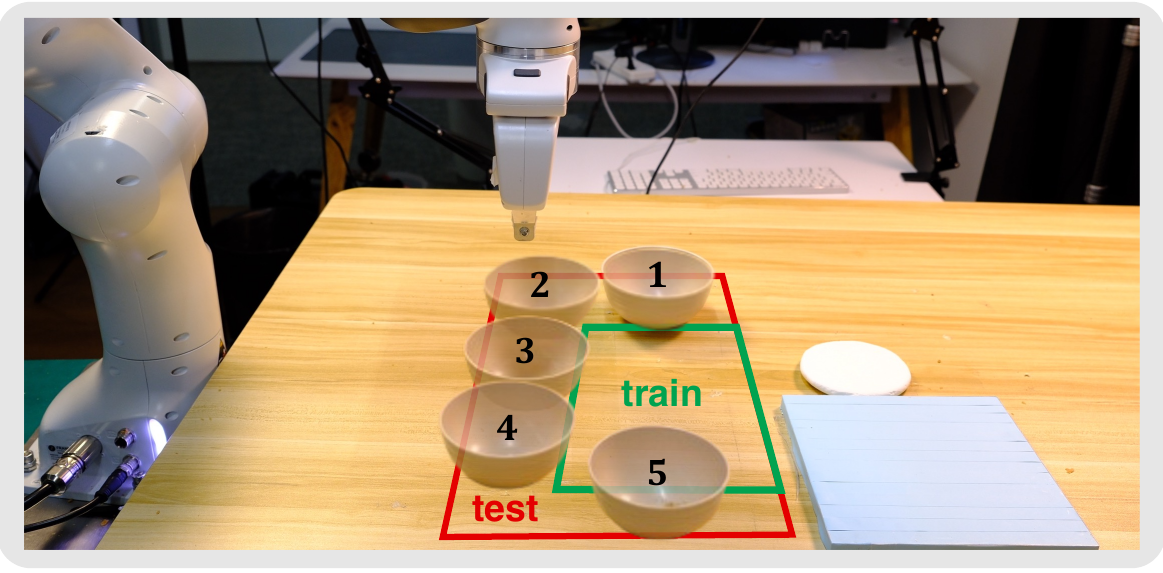}
\resizebox{0.4\textwidth}{!}{%
\vspace{0.2in}
\begin{tabular}{l|cccccc|ccc}
\toprule

  Spatial Generalization &  1 & 2 & 3 & 4 & 5\\

\midrule
Diffusion Policy & \textcolor{gred}{\XSolidBrush} & \textcolor{gred}{\XSolidBrush} & \textcolor{gred}{\XSolidBrush} & \textcolor{gred}{\XSolidBrush} & \textcolor{gred}{\XSolidBrush} \\
  Diffusion Policy (Depth) & \textcolor{gred}{\XSolidBrush} & \textcolor{gred}{\XSolidBrush} & \textcolor{gred}{\XSolidBrush} & \textcolor{gred}{\XSolidBrush} & \textcolor{gred}{\XSolidBrush} \\
  \textbf{\ours}  & \textcolor{gred}{\XSolidBrush} & \textcolor{ggreen}{\Checkmark} & \textcolor{ggreen}{\Checkmark} & \textcolor{ggreen}{\Checkmark} & \textcolor{ggreen}{\Checkmark} \\
\bottomrule
\end{tabular}}
\vspace{-0.1in}
\end{table}

\textbf{Appearance generalization.} \ours is designed to process point clouds without color information, inherently enabling it to generalize across various appearances effectively. As demonstrated in Table~\ref{table: appearance gen}, \ours consistently exhibits successful generalization to cubes of differing colors, while baseline methods could not achieve. It is noteworthy that the depth-based diffusion policy also does not incorporate color as input. However, due to its lower accuracy on the training object, the ability to generalize is also limited.

One solution to improve the appearance generalization ability of image-based methods is applying strong data augmentation during training~\citep{hansen2022lfs,hansen2021svea}, which however could impede the learning process \citep{yuan2022pieg,hansen2021svea}. More importantly, the primary objective of this work is to demonstrate that \ours, even without the aid of any data augmentation, can effectively generalize, thereby underscoring the potential of 3D representations in real robot learning.

\begin{table}[htbp]
\centering
\vspace{-0.05in}
\caption{\textbf{Appearance generalization on Drill.} Algorithms are trained with the green cube only and evaluated on 5 different colored cubes. Each color is evaluated with one trial.}
\label{table: appearance gen}
\vspace{-0.05in}
\includegraphics[width=0.48\textwidth]{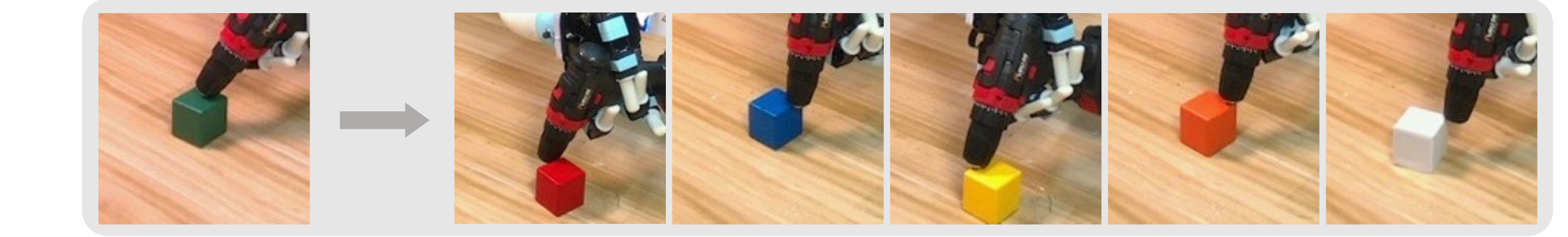}
\resizebox{0.4\textwidth}{!}{%

\begin{tabular}{l|ccccccccc}
\toprule

Apperance Generalization (\textcolor{ggreen}{$\blacksquare$}) & \textcolor{red}{$\blacksquare$}  & \textcolor{blue}{$\blacksquare$}  & \textcolor{yellow}{$\blacksquare$}    & \textcolor{orange}{$\blacksquare$}  & \textcolor{grey}{$\blacksquare$}  \\

\midrule
  Diffusion Policy & \textcolor{gred}{\XSolidBrush} & \textcolor{gred}{\XSolidBrush} & \textcolor{gred}{\XSolidBrush} & \textcolor{gred}{\XSolidBrush} & \textcolor{gred}{\XSolidBrush} \\
  Diffusion Policy (Depth) & \textcolor{gred}{\XSolidBrush} & \textcolor{gred}{\XSolidBrush} & \textcolor{gred}{\XSolidBrush} & \textcolor{gred}{\XSolidBrush} & \textcolor{gred}{\XSolidBrush} \\
  \textbf{\ours}  & \textcolor{ggreen}{\Checkmark} & \textcolor{ggreen}{\Checkmark} & \textcolor{ggreen}{\Checkmark} & \textcolor{ggreen}{\Checkmark} & \textcolor{ggreen}{\Checkmark} \\
\bottomrule
\end{tabular}}
\vspace{-0.1in}
\end{table}

\textbf{Instance generalization.} Achieving generalization across diverse instances, which vary in shape, size, and appearance, presents a significantly greater challenge compared to mere appearance generalization. In Table~\ref{table: instance gen}, we demonstrate that \ours effectively manages a wide range of everyday objects. This success can be primarily attributed to the inherent characteristics of point clouds. Specifically, the use of point clouds allows for policies that are less prone to confusion, particularly when these point clouds are downsampled. This feature significantly enhances the model's ability to adapt to varied instances.

\begin{table}[htbp]
\centering
\vspace{-0.05in}
\caption{\textbf{Instance generalization on Drill.} We replace the cube used in Drill with five objects in varied sizes from our daily life. Each instance is evaluated with one trial.}
\vspace{-0.05in}
\label{table: instance gen}
\includegraphics[width=0.49\textwidth]{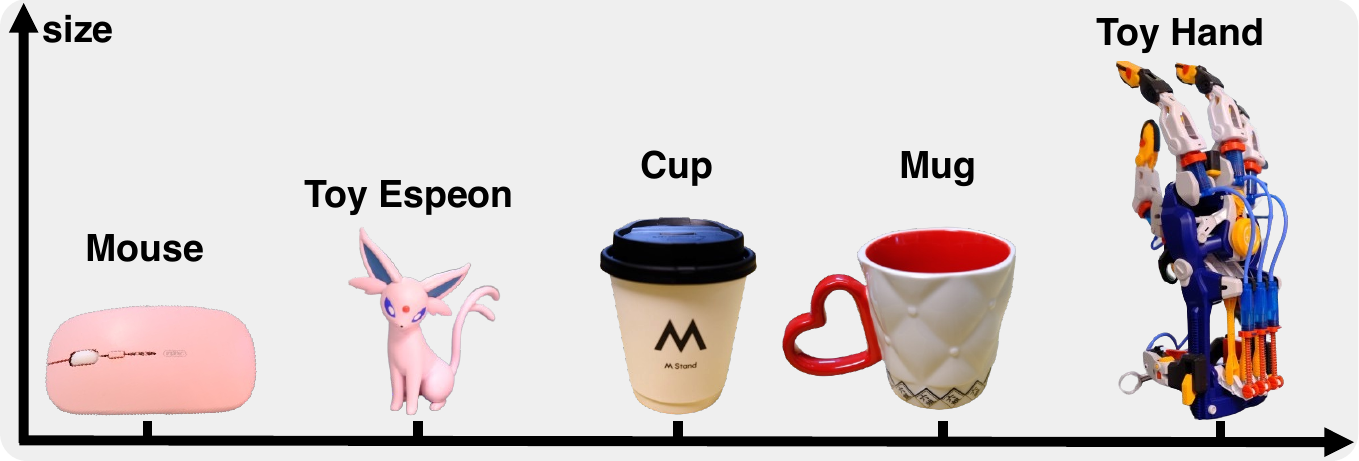}
\resizebox{0.49\textwidth}{!}{%

\begin{tabular}{l|ccccccccc}
\toprule

Instance Generalization & Mouse & Espeon   & Cup    & Mug & Hand  \\

\midrule
  Diffusion Policy & \textcolor{gred}{\XSolidBrush} & \textcolor{gred}{\XSolidBrush} & \textcolor{gred}{\XSolidBrush} & \textcolor{gred}{\XSolidBrush} & \textcolor{ggreen}{\Checkmark} \\
  Diffusion Policy (Depth) & \textcolor{gred}{\XSolidBrush} & \textcolor{gred}{\XSolidBrush} & \textcolor{ggreen}{\Checkmark} & \textcolor{gred}{\XSolidBrush} & \textcolor{gred}{\XSolidBrush} \\
  \textbf{\ours}  & \textcolor{ggreen}{\Checkmark} & \textcolor{ggreen}{\Checkmark} & \textcolor{ggreen}{\Checkmark} & \textcolor{ggreen}{\Checkmark} & \textcolor{ggreen}{\Checkmark} \\
\bottomrule
\end{tabular}}
\end{table}

\textbf{View generalization.} Generalizing image-based methods across different views is notably challenging~\citep{yang2023movie}, and acquiring training data from multiple views can be time-consuming and costly~\citep{ze2023gnfactor,shen2023distilled}. We demonstrate in Table~\ref{table: view gen} that \ours effectively addresses this generalization problem when the camera views are altered slightly. It is important to note that since the camera view is altered, we manually \revise{transform the point clouds} and adjust the cropped space of the point clouds. \revise{Accurate transformation isn't necessary due to the robustness of our network. However, it is crucial to acknowledge that while the network can generalize across minor variations in camera views, significant changes might be hard to handle.}


\begin{table}[htbp]
\vspace{-0.1in}
\centering
\caption{\textbf{View generalization on Roll-Up.} Each view is evaluated with one trial.}
\vspace{-0.05in}
\label{table: view gen}
   \includegraphics[width=0.48\textwidth]{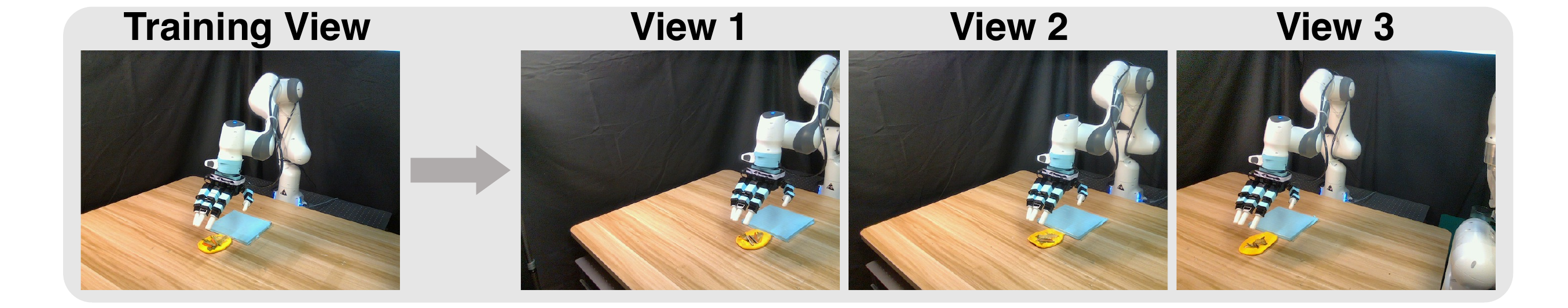}
\resizebox{0.4\textwidth}{!}{%
\begin{tabular}{l|cccccc|ccc}
\toprule

View Generalization & View 1  & View 2  & View 3   \\

\midrule
  Diffusion Policy & \textcolor{gred}{\XSolidBrush} & \textcolor{gred}{\XSolidBrush} & \textcolor{gred}{\XSolidBrush} \\
  Diffusion Policy (Depth) & \textcolor{gred}{\XSolidBrush} & \textcolor{gred}{\XSolidBrush} & \textcolor{gred}{\XSolidBrush} \\
  \textbf{\ours}  & \textcolor{ggreen}{\Checkmark} & \textcolor{ggreen}{\Checkmark} & \textcolor{ggreen}{\Checkmark}  \\
  
\bottomrule
\end{tabular}}
\vspace{-0.1in}
\end{table}

\revise{
\textbf{Cluttered Scenes}. Despite the simplicity of the DP3 Encoder, we demonstrate that DP3 is capable of handling tasks in complex real-world cluttered environments. To illustrate this, we design a pick \& place task (i.e. pick the cube and place it in the bowl) set in cluttered scenes using a gripper and collect 50 demonstrations for training. The results presented in Table~\ref{table: ablation cluttered scenes} show that DP3 solves the task with a high success rate. Aligning with our simulation experiments, DP3 equipped with PointNeXt fails to address the task. Meanwhile, DP3 using color point clouds as input performs comparably to the original DP3 when picking the training yellow cube, yet it struggles with other colored cubes and diverse objects. This demonstrates the effectiveness and generalization ability of DP3 in complex scenes.
}

\begin{table}[htbp]
\centering
\vspace{-0.1in}
\caption{\revise{\textbf{Results in cluttered scenes.} Each algorithm is evaluated with 10 trials in the training color. Each out-of-domain color and object are evaluated with one trial.}}
\label{table: ablation cluttered scenes}
\vspace{-0.05in}
\includegraphics[width=0.48\textwidth]{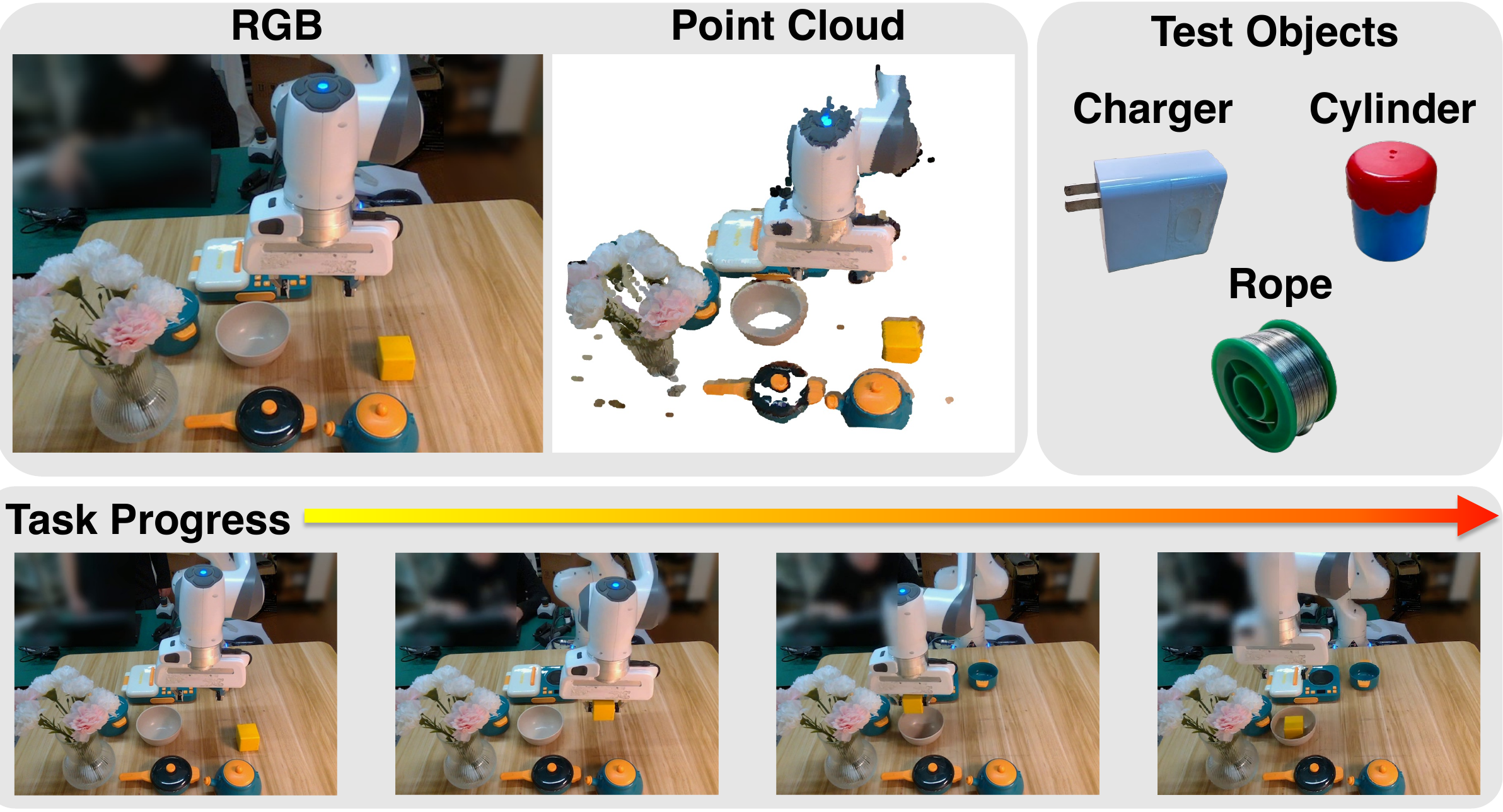}

\resizebox{0.49\textwidth}{!}{%
\begin{tabular}{l|ccccccccc}
\toprule

  Cluttered Scenes  & Diffusion Policy & DP3 w/ PointNeXt  & DP3 w/ color & DP3 \\
\midrule

  Success Rate &  \cc{60} & \cc{0} & \ccbf{80}   & \ccbf{80}\\

\midrule
\end{tabular}}

\resizebox{0.49\textwidth}{!}{%
\begin{tabular}{l|ccc|cccccc}
\midrule

  Train with \textcolor{yellow}{$\blacksquare$} Cube & \textcolor{red}{$\blacksquare$} & \textcolor{blue}{$\blacksquare$} & \textcolor{ggreen}{$\blacksquare$} & Charger & Cylinder & Rope \\
\midrule
 DP3 w/ color & \textcolor{gred}{\XSolidBrush} &
\textcolor{gred}{\XSolidBrush} &
\textcolor{gred}{\XSolidBrush} &\textcolor{gred}{\XSolidBrush} &
\textcolor{gred}{\XSolidBrush} &
\textcolor{gred}{\XSolidBrush} \\
DP3  &  \textcolor{ggreen}{\Checkmark} & \textcolor{ggreen}{\Checkmark} & \textcolor{ggreen}{\Checkmark} &  \textcolor{ggreen}{\Checkmark} & \textcolor{ggreen}{\Checkmark} & \textcolor{ggreen}{\Checkmark} \\

\bottomrule
\end{tabular}}

\end{table}

\subsection{\revise{Observation on Deployment Safety}}
\label{section: safe deployment}
In our real-world experiments, we observe that image-based and depth-based diffusion policies often deliver unpredictable behaviors in real-world experiments, which necessitates human termination to ensure robot safety. We define this situation as \textit{safety violation} and compute the safety violation rate in our main real-world experiments, shown in Table~\ref{table: safety violation}. Interestingly and surprisingly, we find that \ours rarely violates the safety, showing that \ours is a practical and hardware-friendly method for real robot learning. \revise{An intuitive explanation is that since the robots operate in 3D space, directly observing 3D information helps avoid collision. It is important to note that our assessment of safety is primarily qualitative. We intend to explore a more theoretical understanding of this observation in our future work.}

\begin{table}[htbp]
\centering
\caption{\textbf{Safety violation rate.} While conducting the main real-world experiments, we also count the times of safety violation and compute the rate.}
\label{table: safety violation}
\vspace{-0.05in}
\includegraphics[width=0.5\textwidth]{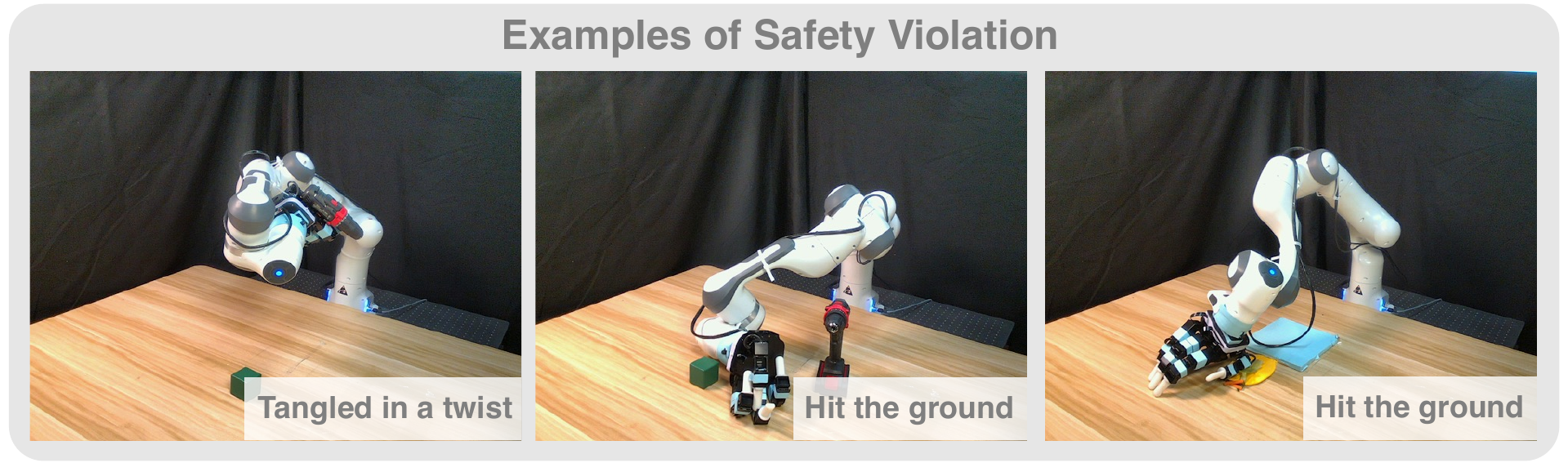}

\resizebox{0.49\textwidth}{!}{%
\begin{tabular}{l|cccc|ccccc}
\toprule

  Safety Violation Rate $\downarrow$ & Roll-Up & Dumpling  & Drill & Pour & \textbf{Average} \\

\midrule

Diffusion Policy &\cc{90} & \cc{20} & \cc{20} & \cc{0} & \cc{32.5}\\
Diffusion Policy (Depth) & \cc{20} & \cc{30} & \cc{30} & \cc{20} & \cc{25.0}\\
\textbf{\ours} & \ccbf{0} & \ccbf{0} & \ccbf{0} &\ccbf{0} & \ccbf{0.0}\\

\bottomrule
\end{tabular}}
\vspace{-0.15in}
\end{table}

\section{Conclusion}
In this work, we introduce \oursfull, an efficient visual imitation learning algorithm, adept at managing a wide range of robotic tasks in both simulated and real-world environments with only a small set of demonstrations. The essence of \ours lies in its integration of carefully designed 3D representations with the expressiveness of diffusion policies. Across 72 simulated tasks,  \ours outperforms its 2D counterpart by a relative margin of \revise{$24.2\%$}.  In real-world scenarios, \ours shows high accuracy in executing complex manipulations of deformable objects using the Allegro hand. More importantly, we demonstrate that \ours possesses robust generalization capabilities across various aspects and causes fewer safety violations in real-world scenarios.

\noindent\textbf{Limitations.} Though we have developed an efficient architecture, the optimal 3D representation for control is still yet discovered. Besides, this work does not delve into tasks with extremely long horizons, which remains for future exploration.

\section*{Acknowledgement}
We would like to thank Zhecheng Yuan, Chen Wang, Cheng Lu, and Jianfei Chen for their helpful discussions. This work is supported by National Key R\&D Program of China (2022ZD0161700).


\bibliographystyle{plainnat}
\bibliography{main}

\clearpage
\newpage
\onecolumn
\begin{appendix}
\subsection{Implementation Details}
\ours mainly consists of two parts: perception and decision. We now detail the implementation details of each part as follows. The official implementation of DP3 is available on \url{https://github.com/YanjieZe/3D-Diffusion-Policy}.

\noindent\textbf{Perception.} The input of \ours includes the visual observation and the robot pose. The visual observation is a point cloud without colors, downsampled from the raw point cloud using Farthest Point Sampling (FPS). We use 512 or 1024 in all the simulated and real-world tasks. DP3 encodes the point cloud into a compact representation with our designed DP3 Encoder. We provide a simple PyTorch implementation of our DP3 Encoder as follows:
\begin{lstlisting}[language=Python, frame=none, basicstyle=\small\ttfamily, commentstyle=\color{ourblue}\small\ttfamily,columns=fullflexible, breaklines=true, postbreak=\mbox{\textcolor{red}{$\hookrightarrow$}\space}, escapeinside={(*}{*)}]
class DP3Encoder(nn.Module):
    def __init__(self, channels=3):
        # We only use xyz (channels=3) in this work
        # while our encoder also works for xyzrgb (channels=6) in our experiments
        self.mlp = nn.Sequential(
                nn.Linear(channels, 64), nn.LayerNorm(64), nn.ReLU(),
                nn.Linear(64, 128), nn.LayerNorm(128), nn.ReLU(),
                nn.Linear(128, 256), nn.LayerNorm(256), nn.ReLU())
        self.projection = nn.Sequential(nn.Linear(256, 64), nn.LayerNorm(64))
    
    def forward(self, x):
        # x: B, N, 3
        x = self.mlp(x) # B, N, 256
        x = torch.max(x, 1)[0] # B, 256
        x = self.projection(x) # B, 64
        return x
\end{lstlisting}
The robot poses are also processed by an MLP network described as follows:
\begin{lstlisting}[language=Python, frame=none, basicstyle=\small\ttfamily, commentstyle=\color{ourblue}\small\ttfamily,columns=fullflexible, breaklines=true, postbreak=\mbox{\textcolor{red}{$\hookrightarrow$}\space}, escapeinside={(*}{*)}]
# DimRobo is the dimension of the robot poses.
Sequential(
  (0): Linear(in_features=DimRobo, out_features=64, bias=True)
  (1): ReLU()
  (2): Linear(in_features=64, out_features=64, bias=True))
\end{lstlisting}
The representations encoded from point clouds and robot poses are concatenated into one representation of dimension 128.  Afterward, the decision backbone generates actions conditioning on this representation.

\noindent\textbf{Decision.} The decision backbone is a convolutional network-based diffusion policy, which transforms random Gaussian noise into a coherent sequence of actions. For implementation, we utilize the official PyTorch framework available from \citep{chi2023diffusion_policy}. In practice, the model is designed to predict a series of $H$ actions based on $N_{obs}$ observed timesteps, but it will only execute the last $N_{act}$ actions during inference. We set $H=4, N_{obs}=2, N_{act}=3$ for \ours and diffusion-based baselines.

The original Diffusion Policy typically employs a longer horizon, primarily due to the denser nature of the timesteps in their tasks.  In Table~\ref{table:prediction horizon}, we show that there is no significant difference between a short horizon and a long horizon for our tasks. Moreover, considering the potential for sudden disruptions in real-world robotic operations, we choose to employ a shorter horizon.

\noindent\textbf{Normalization.} We scale the min and max of each action dimension and each observation dimension to $[-1, 1]$ independently. Normalizing the actions to $[-1,1]$ is a must for the prediction of DDPM and DDIM since they would clip the prediction to $[-1,1]$ for stability. 

\subsection{Task Suite}

\noindent\textbf{Simulated tasks.} We collect diverse simulated tasks to systematically evaluate imitation learning algorithms. Our collected tasks mainly focus on robotic manipulation, including   Adroit~\citep{rajeswaran2017dapg}, Bi-DexHands~\citep{chen2022bidexhands}, DexArt~\citep{bao2023dexart}, DexDeform~\citep{li2023dexdeform}, DexMV~\citep{qin2022dexmv}, HORA~\citep{qi2023hora}, and MetaWorld~\citep{yu2020metaworld}. The full task names could be seen in Table~\ref{table: simulation results}. We add the support for 3D modality in these tasks when the 3D modality is not available originally.

\noindent\textbf{Real-world tasks.} The episode length for our real-world tasks is not fixed. Average episode lengths for demonstrations of each task are listed as follows: (1) 79.9 for Roll-Up; (2) 113.5 for Dumpling; (3) 71.4 for Drill; and (4) 83.6 for Pour. During the evaluation of the policy, we stop the robot when we find (1) the policy finishes the task; (2) the policy can not successfully handle the task; and (3) the policy makes behaviors that are harmful to the hardware. Our real-world setup and everyday objects used in our tasks are shown in Figure~\ref{fig:real world setup}. The randomization in each task is shown in Figure~\ref{fig:real task randomness}. For Roll-Up and Dumpling, the plasticine's shape and the appearance of the objects placed upon the plasticine are randomized. For Drill and Pour, the variations come from the random positions of the cube, drill, and bowl.

\subsection{More Simulation Experiments}
\label{appendix: full simulation results}

\noindent\textbf{Simulation results for each task.} We give the simulation results for each task in Table~\ref{table: simulation results}, which is supplementary to Table~\ref{table:simulation simplified} in our main paper. We report average success rates across 3 seeds. For HORA,  we report the normalized returns since this task is doing in-hand rotation and is not measured by success rates.

\noindent\textbf{Success rates for experts.} In our simulated tasks, we apply Reinforcement Learning (RL)-trained agents to generate demonstrations. These expert policies are rigorously evaluated over 200 episodes, and their success rates are detailed in Table~\ref{table: expert SR}. For MetaWorld tasks, we present results from scripted policies.

\noindent\textbf{Choice of prediction horizon.} DP3 applies a short action prediction and execution horizon $H=4, N_{act}=3$, and so does the baseline Diffusion Policy. This is mainly designed for the generality of \ours in complex tasks and real robot tasks, where the environment would be changed by human intervention and the policy needs to switch action immediately. As shown in Table~\ref{table:prediction horizon}, a shortened prediction horizon is competitive with a longer one.

\begin{table}[htbp]
\centering
\vspace{-0.1in}
\caption{\textbf{Ablation on prediction horizon.} In this work, \ours and Diffusion Policy uses a prediction horizon $H=4, N_{act}=3$. We test $H=16, N_{act}=8$ originally used in \citep{chi2023diffusion_policy} for both methods.}
\label{table:prediction horizon}
\vspace{-0.05in}
\resizebox{0.65\textwidth}{!}{
\begin{tabular}{l|cccccc|ccc}
\toprule

  Algorithm & H & D  & P & A & DA & SP & \textbf{Average} \\

\midrule

\ours & \dd{100}{0} & \dd{62}{4} & \dd{43}{6} & \dd{99}{1} & \dd{69}{4} & \dd{97}{4} & $78.3$\\

w/ long horizon & \dd{100}{0} & \dd{64}{5} & \dd{46}{3} & \dd{99}{1} & \dd{75}{3} & \dd{85}{14} & $78.2$ \\
 
 \midrule
 
Diffusion Policy &	\dd{48}{17} & \dd{50}{5} & \dd{25}{4}	& \dd{15}{1} & \dd{43}{7} & 	\dd{63}{3} & $40.7$\\

w/ long horizon & \dd{68}{11} & \dd{44}{4} & \dd{16}{2} & \dd{12}{3} & \dd{14}{1} & \dd{44}{5} & $33.0$ \\

\bottomrule
\end{tabular}}
\end{table}

\subsection{Simple DP3}

\begin{wrapfigure}{r}{0.35\textwidth}
\vspace{-0.35in}
    \centering
     \includegraphics[width=0.35\textwidth]{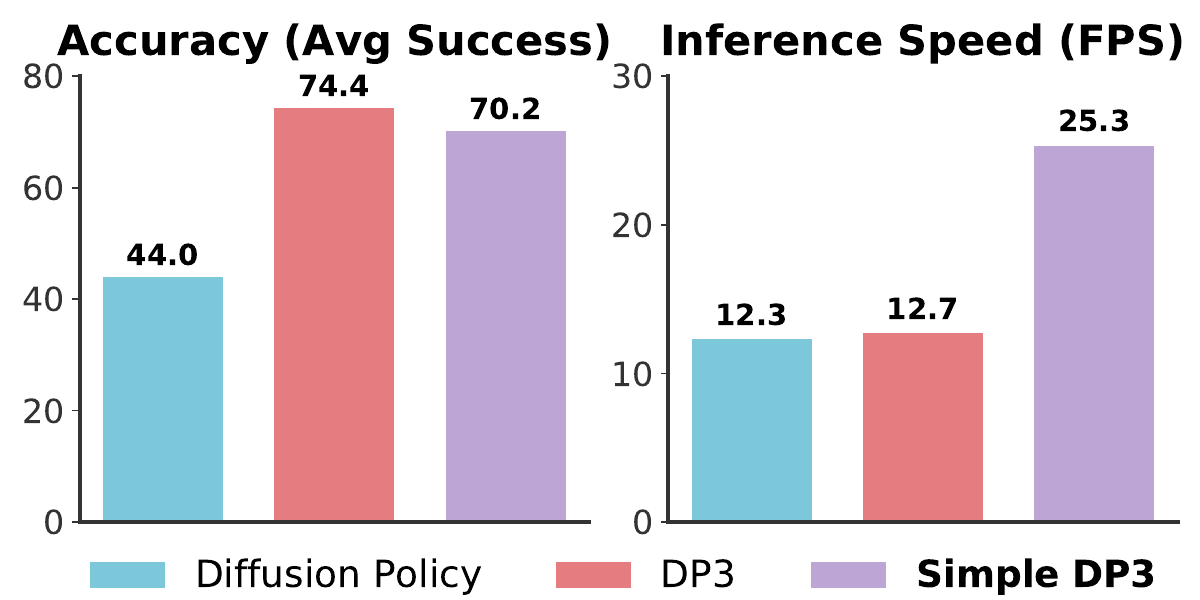}
\end{wrapfigure}

To enhance the applicability of DP3 in real-world robot learning, we simplify the policy backbone of DP3, which is identified as one critical factor that impacts inference speed. The refined version, dubbed \textbf{\textit{Simple DP3}}, offers 2x inference speed while maintaining high accuracy, as shown in Table~\ref{tab:summary simple DP3}. The efficiency stems from removing the redundant components in the UNet backbone. The implementation of Simple DP3 is available on \url{https://github.com/YanjieZe/3D-Diffusion-Policy}.

\begin{table}[htbp]
    \centering
     \caption{\textbf{Results of Simple DP3.} Compared to DP3, Simple DP3 achieves nearly 2x inference speed without losing much accuracy. Full evaluation results are given in Table~\ref{table: simple DP3}.}
    \label{tab:summary simple DP3}
    \vspace{-0.05in}
    \begin{tabular}{l|ccc}
    \toprule
         Algorithm & Diffusion Policy & DP3  & \textbf{Simple DP3} \\
         \midrule
        Inference Speed (FPS) & $12.3$ & $12.7$ & $25.3$ ($\uparrow \mathbf{99}\%$) \\
        Accuracy (Avg Success) & $44.0$ &  $74.4$ & $70.2$ ($\downarrow \mathbf{6}\%$) \\
        \bottomrule
    \end{tabular}

\end{table}

\begin{table*}[htbp]
\centering
\vspace{-0.2in}
\caption{\textbf{Full evaluation results of Simple DP3.} We evaluate Simple DP3 on 10 tasks and compare it with DP3 and find that Simple DP3 could achieve results very competitive to DP3.}
\vspace{-0.05in}
\label{table: simple DP3}
\resizebox{0.9\textwidth}{!}{%
\begin{tabular}{l|ccc|ccc|cccc|cc}
\toprule

    & \multicolumn{3}{c|}{Adroit} & \multicolumn{3}{c|}{MetaWorld} & \multicolumn{4}{c|}{DexArt} & \\
  Algorithm $\backslash$  Task & Hammer & Door  & Pen & Assembly & Disassemble & Stick-Push & Laptop & Faucet & Toilet & Bucket & \textbf{Average} \\

\midrule

\textbf{\ours} & \ddbf{100}{0} & \ddbf{62}{4} & \ddbf{43}{6} & \ddbf{99}{1} & \ddbf{69}{4} & \ddbf{97}{4} & \ddbf{83}{1} & \ddbf{63}{2} & \ddbf{82}{4} & \ddbf{46}{2} & \ccbf{74.4}\\

Diffusion Policy &	\dd{48}{17} & \dd{50}{5} & \dd{25}{4}	& \dd{15}{1} & \dd{43}{7} & 	\dd{63}{3} & \dd{69}{4} & \dd{23}{8}  & \dd{58}{2} & \ddbf{46}{1} & $44.0$ \\

\textbf{Simple DP3} & \ddbf{100}{0} & \dd{58}{4} & \ddbf{46}{5} & \dd{79}{1} & \dd{50}{3} & \ddbf{97}{5} & \ddbf{84}{2} & \ddbf{63}{3} & \dd{81}{6} & \dd{44}{6} & $70.2$ \\

\bottomrule
\end{tabular}}
\vspace{-0.2in}
\end{table*}

\begin{table*}[htbp]
\centering
\caption{\textbf{Main results on 72 simulation tasks.} Results for each task are provided in this table. A summary across domains is shown in Table~\ref{table:simulation simplified}. }
\label{table: simulation results}
\begin{flushleft}
\resizebox{0.9\textwidth}{!}{%
\begin{tabular}{l|ccc|ccccccc}
\toprule
& \multicolumn{3}{c|}{\textbf{Adroit}~\citep{rajeswaran2017dapg}} & \multicolumn{6}{c}{\textbf{Bi-DexHands}~\citep{chen2022bidexhands}} \\

 Alg $\backslash$ Task & Hammer  & Door & Pen & Block Stack & Bottle Cap & Door Open Outward & Grasp And Place & Hand Over & Scissors\\
 
\midrule

\ours &  \ddbf{100}{0} & \ddbf{62}{4} & \ddbf{43}{6} & \ddbf{24}{15} & \ddbf{83}{10} & \ddbf{100}{0} & \ddbf{69}{22} & \ddbf{45}{8} & \ddbf{100}{0} \\
 Diffusion Policy &   \dd{45}{5} & \dd{37}{2} & \dd{13}{2} & \dd{4}{4} & \dd{61}{5}  & \ddbf{100}{0} & \dd{65}{9} & \dd{38}{0} & \ddbf{100}{0} \\
\bottomrule
\end{tabular}}

\resizebox{0.9\textwidth}{!}{%
\begin{tabular}{l|cccc|cccccc|cc|cc}
\toprule
& \multicolumn{4}{c|}{\textbf{DexArt}~\citep{bao2023dexart}} & \multicolumn{6}{c|}{\textbf{DexDeform}~\citep{li2023dexdeform}} & \multicolumn{2}{c|}{\textbf{DexMV}~\citep{qin2022dexmv}} & \textbf{HORA}~\citep{qi2023hora}\\

 Alg $\backslash$ Task & Laptop & Faucet & Toilet & Bucket & Rope & Bun & Dumpling & Wrap & Flip & Folding & Pour & Place Inside & Rotation\\
 
\midrule

\ours & \ddbf{83}{1}  & \ddbf{63}{2} & \ddbf{82}{4} & \ddbf{46}{2} & \dd{93}{2} & \dd{70}{9} & \ddbf{92}{0} & \ddbf{94}{0} & \dd{97}{1} &  \dd{81}{2} & \ddbf{99}{2} & \ddbf{100}{0}  & \ddbf{71}{31}   \\

 Diffusion Policy & \dd{69}{4} & \dd{23}{8} & \dd{58}{2} & \ddbf{46}{1} & \ddbf{97}{0} & \ddbf{76}{4} &  \ddbf{92}{0} & \dd{91}{0} & \ddbf{99}{0} & \ddbf{88}{1} & \dd{90}{2} & \ddbf{100}{0} & \dd{49}{11} \\
\bottomrule
\end{tabular}}

\resizebox{0.9\textwidth}{!}{%
\begin{tabular}{l|cccccccc}
\toprule
& \multicolumn{7}{c}{\textbf{Meta-World~\citep{yu2020metaworld} (Easy)}} \\

 Alg $\backslash$ Task & Button Press & Button Press Topdown & Button Press Topdown Wall & Button Press Wall & Coffee Button & Dial Turn & Door Close \\
\midrule

\ours & \ddbf{100}{0} & \ddbf{100}{0} & \ddbf{99}{2} & \ddbf{99}{1} & \ddbf{100}{0} & \ddbf{66}{1} &  \ddbf{100}{0}     \\
 Diffusion Policy & \dd{99}{1} & \dd{98}{1} & \dd{96}{3} & \dd{97}{3} & \dd{99}{1} & \dd{63}{10} & \ddbf{100}{0}  \\

\bottomrule
\end{tabular}}

\resizebox{0.9\textwidth}{!}{%
\begin{tabular}{l|ccccccccccc}
\toprule
& \multicolumn{9}{c}{\textbf{Meta-World (Easy)}} \\

 Alg $\backslash$ Task & Door Lock & Door Open & Door Unlock & Drawer Close & Drawer Open & Faucet Close & Faucet Open & Handle Press & Handle Pull\\
\midrule
\ours & \ddbf{98}{2} & \ddbf{99}{1} & \ddbf{100}{0} &   \ddbf{100}{0} & \ddbf{100}{0} & \ddbf{100}{0} & \ddbf{100}{0} & \ddbf{100}{0} &  \ddbf{53}{11} &     \\
 Diffusion Policy &  \dd{86}{8} & \dd{98}{3}& \dd{98}{3}& \ddbf{100}{0}& \dd{93}{3}& \ddbf{100}{0}& \ddbf{100}{0}& \dd{81}{4}& \dd{27}{22} \\

\bottomrule
\end{tabular}}

\resizebox{0.9\textwidth}{!}{%
\begin{tabular}{l|ccccccccccc}
\toprule
& \multicolumn{8}{c}{\textbf{Meta-World (Easy)}} \\

 Alg $\backslash$ Task & Handle Press Side & Handle Pull Side & Lever Pull & Plate Slide & Plate Slide Back & Plate Slide Back Side & Plate Slide Side & Reach\\
\midrule

\ours & \ddbf{100}{0} & \ddbf{85}{3} & \ddbf{79}{8} & \ddbf{100}{1} & \ddbf{99}{0} & \ddbf{100}{0} & \ddbf{100}{0} & \ddbf{24}{1}\\

Diffusion Policy & \ddbf{100}{0}& \dd{23}{17}& \dd{	49}{5}& \dd{	83}{4}& \ddbf{	99}{0}& \ddbf{	100}{0}& \ddbf{	100}{0}& \dd{	18}{2}\\

\bottomrule
\end{tabular}}

\resizebox{0.9\textwidth}{!}{%
\begin{tabular}{l|cccc|ccccccc}
\toprule
& \multicolumn{4}{c|}{\textbf{Meta-World (Easy)}} & \multicolumn{5}{c}{\textbf{Meta-World (Medium)}} \\

 Alg $\backslash$ Task & Reach Wall & Window Close & Window Open & Peg Unplug Side & Basketball & Bin Picking & Box Close & Coffee Pull & Coffee Push\\
\midrule

\ours & \ddbf{68}{3}& \ddbf{100}{0}& \ddbf{100}{0}& \ddbf{75}{5} & \ddbf{98}{2} & \ddbf{34}{30} & \ddbf{42}{3} & \ddbf{87}{3} & \ddbf{94}{3}\\
 Diffusion Policy &\dd{59}{7}&\dd{100}{0}&\dd{100}{0}&\dd{74}{3} &\dd{85}{6}&\dd{15}{4}&\dd{30}{5}&\dd{34}{7}&\dd{67}{4} \\

\bottomrule
\end{tabular}}

\resizebox{0.9\textwidth}{!}{%
\begin{tabular}{l|cccccc|ccccc}
\toprule
& \multicolumn{6}{c|}{\textbf{Meta-World (Medium)}} & \multicolumn{4}{c}{\textbf{Meta-World (Hard)}} \\

 Alg $\backslash$ Task & Hammer & Peg Insert Side & Push Wall & Soccer & Sweep & Sweep Into & Assembly & Hand Insert & Pick Out of Hole & Pick Place\\
\midrule

\ours & \ddbf{76}{4}& \ddbf{69}{7}& \ddbf{49}{8}& \ddbf{18}{3}& \ddbf{96}{3}& \ddbf{15}{5}& \ddbf{99}{1}& \ddbf{14}{4}& \ddbf{14}{9}& \ddbf{12}{4}\\

 Diffusion Policy & \dd{15}{6}& \dd{34}{7}& \dd{20}{3}& \dd{14}{4}& \dd{18}{8}& \dd{10}{4}& \dd{15}{1}& \dd{9}{2}& \dd{0}{0}& \dd{0}{0}\\

\bottomrule
\end{tabular}}

\resizebox{0.9\textwidth}{!}{%
\begin{tabular}{l|cc|ccccc|ccccccc}
\toprule
& \multicolumn{2}{c|}{\textbf{Meta-World (Hard)}} & \multicolumn{5}{c|}{\textbf{Meta-World (Very Hard)}} & 
 \multicolumn{1}{c}{\multirow{2}{*}{\textbf{Average}}} & \\

 Alg $\backslash$ Task & Push & Push Back & Shelf Place & Disassemble & Stick Pull & Stick Push & Pick Place Wall &  \\
\midrule

\ours & \ddbf{51}{3}& \ddbf{0}{0}& \ddbf{17}{10}& \ddbf{69}{4}& \ddbf{27}{8}& \ddbf{97}{4}& \ddbf{35}{8} &  \multicolumn{1}{c}{\ccbf{74.4}}\\

 Diffusion Policy &\dd{30}{3}& \dd{0}{0}& \dd{11}{3}& \dd{43}{7}& \dd{11}{2}& \dd{63}{3}& \dd{5}{1} & \multicolumn{1}{c}{{59.8}}\\

\bottomrule
\end{tabular}}

\end{flushleft}
\end{table*}

\begin{table*}[htbp]
\centering
\caption{\textbf{Success rates of experts on 72 simulation tasks.} We evaluate 200 episodes for each task. For MetaWorld tasks, we evaluate both BAC agents and the script policies provided officially in MetaWorld. For DexDeform tasks, the demonstrations are collected by human teleportation~\citep{li2023dexdeform} and we record the success rates as 100\%.}
\label{table: expert SR}
\vspace{-0.1in}
\begin{flushleft}
\resizebox{0.9\textwidth}{!}{%
\begin{tabular}{l|ccc|ccccccc}
\toprule
& \multicolumn{3}{c|}{\textbf{Adroit}~\citep{rajeswaran2017dapg}} & \multicolumn{6}{c}{\textbf{Bi-DexHands}~\citep{chen2022bidexhands}} \\

 Alg $\backslash$ Task & Hammer  & Door & Pen & Block Stack & Bottle Cap & Door Open Outward & Grasp And Place & Hand Over & Scissors\\
 
\midrule

Expert & \cc{99.0} & \cc{100.0} & \cc{97.0} & \cc{83.5} & \cc{100.0} & \cc{100.0} & \cc{100.0} & \cc{77.0} & \cc{99.5}\\
 
\bottomrule
\end{tabular}}

\resizebox{0.9\textwidth}{!}{%
\begin{tabular}{l|cccc|cccccc|cc|cc}
\toprule
& \multicolumn{4}{c|}{\textbf{DexArt}~\citep{bao2023dexart}} & \multicolumn{6}{c|}{\textbf{DexDeform}~\citep{li2023dexdeform}} & \multicolumn{2}{c|}{\textbf{DexMV}~\citep{qin2022dexmv}} & \textbf{HORA}~\citep{qi2023hora}\\

 Alg $\backslash$ Task & Laptop & Faucet & Toilet & Bucket & Rope & Bun & Dumpling & Wrap & Flip & Folding & Pour & Place Inside & Rotation\\
 
\midrule

Expert & \cc{86.5} & \cc{58.0} & \cc{66.5} & \cc{80.0} & \cc{100.0} & \cc{100.0} &  \cc{100.0} & \cc{100.0} & \cc{100.0} & \cc{100.0} & \cc{88.5} & \cc{64.5} & \cc{80.5}\\
\bottomrule
\end{tabular}}

\resizebox{0.9\textwidth}{!}{%
\begin{tabular}{l|cccccccc}
\toprule
& \multicolumn{7}{c}{\textbf{Meta-World~\citep{yu2020metaworld} (Easy)}} \\

 Alg $\backslash$ Task & Button Press & Button Press Topdown & Button Press Topdown Wall & Button Press Wall & Coffee Button & Dial Turn & Door Close \\
\midrule

Expert & \cc{100.0} & \cc{100.0} & \cc{100.0} & \cc{98.5} & \cc{100.0} & \cc{100.0} & \cc{100.0}\\

\bottomrule
\end{tabular}}

\resizebox{0.9\textwidth}{!}{%
\begin{tabular}{l|ccccccccccc}
\toprule
& \multicolumn{9}{c}{\textbf{Meta-World (Easy)}} \\

 Alg $\backslash$ Task & Door Lock & Door Open & Door Unlock & Drawer Close & Drawer Open & Faucet Close & Faucet Open & Handle Press & Handle Pull\\
\midrule

 Expert & \cc{100.0} & \cc{98.5} & \cc{100.0} & \cc{100.0} &\cc{100.0} & \cc{100.0} & \cc{100.0} & \cc{100.0} & \cc{100.0}\\
 
\bottomrule
\end{tabular}}

\resizebox{0.9\textwidth}{!}{%
\begin{tabular}{l|ccccccccccc}
\toprule
& \multicolumn{8}{c}{\textbf{Meta-World (Easy)}} \\

 Alg $\backslash$ Task & Handle Press Side & Handle Pull Side & Lever Pull & Plate Slide & Plate Slide Back & Plate Slide Back Side & Plate Slide Side & Reach\\
\midrule

Expert & \cc{100.0} & \cc{100.0} & \cc{98.5} & \cc{100.0} & \cc{100.0} & \cc{100.0} & \cc{100.0} & \cc{100.0}\\
\bottomrule
\end{tabular}}

\resizebox{0.9\textwidth}{!}{%
\begin{tabular}{l|cccc|ccccccc}
\toprule
& \multicolumn{4}{c|}{\textbf{Meta-World (Easy)}} & \multicolumn{5}{c}{\textbf{Meta-World (Medium)}} \\

 Alg $\backslash$ Task & Reach Wall & Window Close & Window Open & Peg Unplug Side & Basketball & Bin Picking & Box Close & Coffee Pull & Coffee Push\\
\midrule

 Expert & \cc{100.0} & \cc{100.0} & \cc{100.0} & \cc{99.0} &\cc{100.0} & \cc{97.0} & \cc{90.0} & \cc{100.0} & \cc{100.0}\\
\bottomrule
\end{tabular}}

\resizebox{0.9\textwidth}{!}{%
\begin{tabular}{l|cccccc|ccccc}
\toprule
& \multicolumn{6}{c|}{\textbf{Meta-World (Medium)}} & \multicolumn{4}{c}{\textbf{Meta-World (Hard)}} \\

 Alg $\backslash$ Task & Hammer & Peg Insert Side & Push Wall & Soccer & Sweep & Sweep Into & Assembly & Hand Insert & Pick Out of Hole & Pick Place\\
\midrule

Expert & \cc{100.0} & \cc{92.0} & \cc{100.0} & \cc{90.5} &\cc{100.0} & \cc{90.0} & \cc{100.0} & \cc{100.0} & \cc{100.0} & \cc{100.0}\\
\bottomrule
\end{tabular}}

\resizebox{0.7\textwidth}{!}{%
\begin{tabular}{l|cc|cccccccccccc}
\toprule
& \multicolumn{2}{c|}{\textbf{Meta-World (Hard)}} & \multicolumn{5}{c}{\textbf{Meta-World (Very Hard)}}

  \\

 Alg $\backslash$ Task & Push & Push Back & Shelf Place & Disassemble & Stick Pull & Stick Push & Pick Place Wall  \\
\midrule

 Expert & \cc{100.0} & \cc{0.0} & \cc{99.5} & \cc{92.5} & \cc{95.0} & \cc{100.0} & \cc{99.5}\\
\bottomrule
\end{tabular}}
\end{flushleft}

\end{table*}

\end{appendix}

\end{document}